%% file: main_aistats.tex
\documentclass[twoside]{article}

\usepackage[accepted]{aistats2026}

\usepackage{enumitem}
\setlist[itemize]{noitemsep}

\usepackage{graphicx} %
\usepackage{caption}

\usepackage{algorithm}
\usepackage{algpseudocode}

\input{math_commands.tex}

\usepackage{url}

\usepackage{subcaption}

\graphicspath{ {./images/} }

\usepackage{booktabs}

\newcommand{\yj}[1]{#1}

\usepackage[round]{natbib}
\renewcommand{\bibname}{References}
\renewcommand{\bibsection}{\subsubsection*{\bibname}}

\newcommand{\authornotes}{%
\begingroup%
\renewcommand{\thefootnote}{}%
\footnotetext{\kern-1.8em$^*$Equal contribution. $^\dagger$Work completed prior to joining NVIDIA.}%
\endgroup%
}

\input{preamble}

\begin{document}

\twocolumn[
\aistatstitle{Tractable Uncertainty-Aware Meta-Learning}

\aistatsauthor{ Young-Jin Park$^{*}$ \And Cesar Almecija$^{*}$ \And Apoorva Sharma$^{\dagger}$ \And Navid Azizan}

\aistatsaddress{MIT \And MIT \And NVIDIA \And MIT}

]

\runningauthor{Young-Jin Park, Cesar Almecija, Apoorva Sharma, and Navid Azizan}

\begin{abstract}
Meta-learning is a popular approach for learning new tasks with limited data by leveraging the commonalities among different tasks. 
However, meta-learned models can perform poorly when context data is too limited, or when data is drawn from an out-of-distribution (OoD) task. 
Especially in safety-critical settings, this necessitates an uncertainty-aware approach to meta-learning.
In addition, the often multimodal nature of task distributions can pose unique challenges to meta-learning methods.
To this end, we present \approach{}, a meta-learning method for \emph{regression} that (1) makes probabilistic predictions on in-distribution tasks efficiently, (2) is capable of detecting OoD context data, and (3) handles heterogeneous, multimodal task distributions effectively.
The strength of our framework lies in its solid theoretical basis, enabling \revise{analytically tractable Bayesian inference on a linearized model} for principled uncertainty estimation and robust generalization.
We achieve this by adopting a probabilistic perspective and learning a parametric, tunable task distribution via Bayesian inference on a linearized neural network, leveraging Gaussian process theory.
Moreover, we make our approach computationally tractable by leveraging a low-rank prior covariance learning scheme based on the Fisher Information Matrix.
Our numerical analysis demonstrates that \approach{} quickly adapts to new tasks and remains accurate even in low-data regimes; it effectively detects OoD tasks; and that both of these properties continue to hold for multimodal task distributions.
\end{abstract}

\section{Introduction} \authornotes

Learning to learn is arguably an essential part of natural intelligence but is still an active area of research in artificial intelligence.
\textit{Meta-learning} is a popular approach that aims to enable trained models to perform well on new tasks using limited data from the new task.
It involves first a \textit{meta-training} process, when the model learns useful features from a set of tasks.
Then, at test time, using only a few datapoints (\textit{context data}) from a new, unseen task, the model (1) \textit{adapts} to the new task (i.e., performs \textit{few-shot learning})
and then (2) \textit{infers} by making predictions on new, unseen \textit{query inputs} from the same task.
A popular baseline for meta-learning, which has attracted considerable attention in the past few years, is model-agnostic meta-learning (MAML) \citep{maml}, in which the adaptation process consists of fine-tuning the parameters of the model via gradient descent.

Despite their success, meta-learning methods can struggle in several ways when deployed in challenging real-world scenarios. First, when context data is too limited to fully identify the test-time task, accurate prediction can be challenging. As these predictions can be untrustworthy, this necessitates the development of meta-learning methods that can express uncertainty during adaptation \citep{bayesian_maml, alpaca}. In addition, meta-learning models may not successfully adapt to ``unusual'' tasks, i.e., when test-time context data is drawn from an \textit{out-of-distribution} (OoD) task not well represented in the training dataset \citep{ood_maml, meta_learning_ood}.
Finally, special care has to be taken when learning heterogeneous tasks.
An important example is the case of tasks with a \textit{multimodal} distribution, i.e., when there are no common features shared across all the tasks, but the tasks can be grouped into subsets (modes) in a way that the ones from the same subset share common features \citep{mmaml}.

To address these challenges, we present \approach{} (\approachLong), a meta-learning method that leverages probabilistic tools to overcome the aforementioned issues for \emph{regression} tasks.
Specifically, \approach{} models the true distribution of tasks with a learnable distribution constructed over a linearized neural network and uses analytic Bayesian inference to perform uncertainty-aware adaptation.
Further, we show how \approach{} effectively strikes a balance between learning a rich prior distribution over the weights and maintaining the expressivity of the network. %
Finally, through numerical analysis, we demonstrate that (1) our method allows for efficient probabilistic predictions on in-distribution tasks, (2) it is effective in detecting context data from OoD tasks at test time, and (3) both these findings continue to hold in the multimodal task-distribution setting.

In short, our key contributions are:
\vspace{-0.5em}
\begin{itemize}[leftmargin=10pt]
\item We introduce a meta-learning framework for regression that models the task distribution via Bayesian inference on a linearized neural network (Section~\ref{sec:posterior}). This approach uniquely enables \textbf{analytically tractable posterior inference}, avoiding common (e.g., sample-based) approximations.
\item To make our method scalable for deep networks, we introduce an efficient low-rank parameterization of the prior weight covariance based on the Fisher Information Matrix (FIM), making the approach \textbf{computationally tractable} (Section~\ref{sec:prior_covariance}).
\item The framework is extended to effectively handle \textbf{heterogeneous tasks} by modeling the task distribution as a mixture of Gaussian Processes, allowing it to adapt to different underlying task clusters. (Section~\ref{sec:mixture})
\item \revise{The analytically tractable posterior on the linearized model} yields \textbf{principled uncertainty estimates} that provide superior OoD detection and robust few-shot learning performance, especially in low-data regimes.
\end{itemize}

\section{Related Work}
\label{sec:related_work}
\textbf{Bayesian inference with linearized DNNs.}
Bayesian inference with neural networks is often intractable. Whereas \approach{} linearizes the network to allow for practical Bayesian inference, existing work has used other approximations such as Laplace's method. Closely related to our work, \citet{maddox} have linearized pre-trained networks and performed domain adaptation by conditioning the prior predictive with data from the new task. Our approach leverages a similar adaptation method and demonstrates how the prior distribution can be learned in a meta-learning setup.

\textbf{Meta-learning.}
Probabilistic meta-learning models such as PLATIPUS or BaMAML \citep{bayesian_maml, probabilistic_maml} augment MAML to perform approximate Bayesian inference. These approaches, like ours, learn (during meta-training) and make use of (at test-time) a prior distribution over the weights. In contrast, however, \approach{} \revise{performs analytically tractable Bayesian inference on a linearized model at test-time}.
Therefore, unlike other probabilistic frameworks that estimate the posterior predictive distribution through sampling, our method yields an \emph{analytically tractable} posterior distribution.

ALPaCA is a Bayesian meta-learning algorithm for neural networks, where only the last layer is Bayesian \citep{alpaca}.
This framework yields an exact linear regression that uses as feature map the activations right before the last layer.
Our work can be viewed as a generalization of ALPaCA, in the sense that \approach{} restricted to the last layer matches ALPaCA's approach.
The link between these methods is further discussed in Appendix~\ref{app:link_with_alpaca}.

\revise{\textbf{Neural processes.}
Neural Processes (NPs) \citep{np} are a family of meta-learning methods that parameterize stochastic processes via neural networks. Conditional Neural Processes (CNPs) \citep{cnp} use a permutation-invariant encoder to aggregate context data into a fixed-length latent representation for prediction. Transformer Neural Processes (TNP-D) \citep{tnpd} leverage attention mechanisms to capture richer context dependencies, achieving strong performance across a range of tasks.
However, the encoder-based architecture of NPs maps context sets to a single latent representation, which can make it challenging to handle multimodal task distributions and to distinguish OoD tasks from in-distribution tasks.
In contrast, our framework leverages a mixture-of-GPs formulation on linearized networks, providing analytically tractable per-component marginal likelihoods.
This analytical inference allows for more principled and robust task-level OoD detection.
}

\revise{\textbf{Deep kernel learning.}
Deep Kernel Transfer (DKT) combines deep feature extractors with GP inference by defining a kernel over learned feature outputs \citep{dkt}. While DKT can leverage powerful backbones for strong regression performance, its uncertainty operates at the input level based on distance in feature space, rather than the task level. Our method operates in weight space via the Neural Tangent Kernel, enabling direct evaluation of the prior predictive for task-level uncertainty assessment.}

\textbf{Meta-learning vs.\ fine-tuning.}
A widely used approach for adapting foundation models is fine-tuning, but it can be computationally expensive and struggle when only a small number of labeled examples are available. Meta-learning offers a more principled framework for adapting to families of related tasks, allowing for rapid generalization and greater robustness to domain shifts, particularly in low-data regimes.

A more comprehensive discussion of related work can be found in Appendix~\ref{sec:related_work_appendix}.

\begin{figure*}[t]
    \centering
    \includegraphics[width=0.80\textwidth]{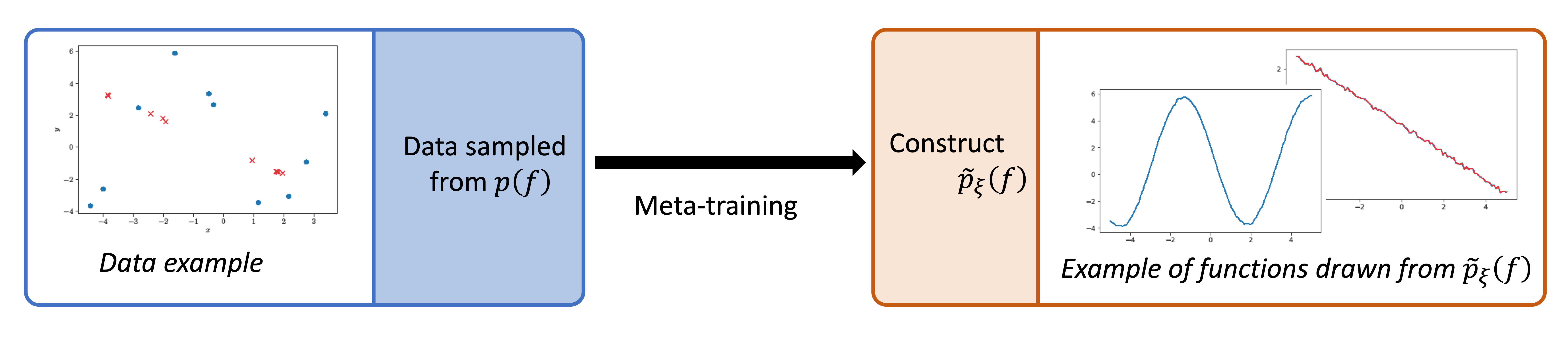}
    \caption{The true task distribution $\dist$ can be multimodal, with multiple task clusters (e.g., lines and sines). \approach{} models $\dist$ with a tunable parametric distribution $\distparam$ via Bayesian linear regression on a linearized neural network.}
    \label{fig:flowchart}
\end{figure*}

\section{Problem Statement}\label{sec:problem_statement}
At test time, the prediction steps are broken down into (1) \textit{adaptation}, that is identifying $f_i$ using $\batchsize$ context datapoints $(\xcontextinput, \ycontextoutput)$ from the task, and (2) \textit{inference}, that is making predictions for $f_i$ on the \textit{query inputs} $\xqueryinput$.
Later the predictions can be compared with the \textit{query ground-truths} $\yqueryoutput$ to estimate the quality of the prediction, for example in terms of mean squared error (MSE).
The meta-training consists in learning valuable features from a \textit{cluster of tasks}, which is a set of similar tasks (e.g., sines with different phases and amplitudes but same frequency), so that at test time the predictions can be accurate on tasks from the same cluster.
We take a probabilistic, functional perspective and represent a cluster by $\dist$, a theoretical distribution over the function space that describes the probability of a task belonging to the cluster.
Learning $\dist$ is appealing, as it allows for performing OoD detection in addition to making predictions. Adaptation amounts to computing the conditional distribution given test context data, and one can obtain an uncertainty metric by evaluating the negative log-likelihood (NLL) of the context data under $\dist$.

Thus, our goal is to construct a parametric, learnable functional distribution $\distparam$ that approaches the theoretical distribution $\dist$, with a structure that allows tractable conditioning and likelihood computation, even in deep learning contexts.
In practice, however, we are not given $\dist$, but only a meta-training dataset $\dataset$ that we assume is sampled from $\dist$: $\dataset=\{ (\xallcontextinput, \yallcontextoutput)\}_{i=1}^{\ntraintasks}$, where $\ntraintasks$ is the number of tasks available during training, \revise{the superscript $i$ indexes each task, and} $(\xallcontextinput, \yallcontextoutput) \sim \task$ is the entire pool of data from which we can draw subsets of context data $(\xcontextinput, \ycontextoutput)$.
Consequently, in the meta-training phase, we aim to optimize $\distparam$ to capture properties of $\dist$, using only the samples in $\dataset$, as illustrated in Figure~\ref{fig:flowchart}.

Once we have $\distparam$, we can evaluate it both in terms of how it performs for few-shot learning (by comparing the predictions with the ground truths in terms of MSE), as well as for OoD detection 
(by measuring how well the NLL of context data serves to classify in-distribution tasks against OoD tasks, measured via the AUC-ROC score). \nopagebreak[4]

\section{Background}
\label{sec:background}
\subsection{Bayesian linear regression and Gaussian Processes}
\label{sec:reglin}
Efficient Bayesian meta-learning requires a tractable inference process at test time. In general, this is only possible analytically in a few cases.
One of them is the Bayesian linear regression with Gaussian noise and a Gaussian prior on the weights. 
Viewing it from a nonparametric, functional approach, this model is equivalent to a Gaussian process (GP) \citep{rasmussen}.

Let $\xinput = (\xsingleinput_1, \dots, \xsingleinput_\batchsize) \in \R^{\revise{\xdim} \times \batchsize}$ be a batch of $\batchsize$ $\xdim$-dimensional inputs, and let $\yvectorizedoutput = (\ysingleoutput_1, \dots, \ysingleoutput_\batchsize) \in \R^{\revise{\ydim} \batchsize}$ be a vectorized batch of $\ydim$-dimensional outputs. In the Bayesian linear regression model, these quantities are related according to
$
    \yvectorizedoutput = \featuremap(\xinput)^\top \paramcorr + \noise \in \R^{\ydim \batchsize}
$
where $\paramcorr \in \R^\pdim$ are the weights of the model, and the inputs are mapped via $\featuremap:\R^{\xdim \times \batchsize} \rightarrow \R^{\pdim \times \ydim \batchsize}$. Notice how this is a generalization of the usual one-dimensional linear regression ($\ydim=1$).

If we assume a Gaussian prior on the weights $\paramcorr \sim \normal(\priormean, \priorcov)$ and a Gaussian noise $\noise \sim \normal(\zeromean, \noisecov)$ with $\noisecov = \stdnoise^2 \mI$, then the model describes a multivariate Gaussian distribution on $\yvectorizedoutput$ for any $\xinput$. Equivalently, this means that this model describes a GP distribution over functions, with mean and covariance function (or kernel)
\begin{align}
\begin{split}
\label{eq:prior_pred_dist}
\meanpriorpred (\xsingleinput_t) & = \featuremap(\xsingleinput_t)^\top \priormean, \\
\covpriorpred  (\xsingleinput_{t_1}, \xsingleinput_{t_2}) & =  \featuremap(\xsingleinput_{t_1})^\top \priorcov \featuremap(\xsingleinput_{t_2}) + \noisecov  \\
& =: \kernel_\priorcov(\xsingleinput_{t_1}, \xsingleinput_{t_2}) + \noisecov .
\end{split}
\end{align}
This GP enables tractable computation of the likelihood of any batch of data $(\xinput, \youtput)$ given this distribution over functions. The structure of this distribution is governed by the feature map $\featuremap$ and the prior over the weights, specified by $\priormean$ and $\priorcov$.

This distribution can also easily be conditioned to perform inference. Given a batch of data $(\xinput, \youtput)$, the posterior predictive distribution is also a GP, with an updated mean and covariance function
\begin{align}
    \label{eq:post_pred_dist}
    \begin{split}
        & \meanpostpred (\xsingleinput_{t_*}) =  \kernel_\priorcov(\xsingleinput_{t_*}, \xinput) \left( \kernel_\priorcov(\xinput, \xinput) + \noisecov \right)^{-1} \youtput, \\
        & \covpostpred (\xsingleinput_{{t_1}_*}, \xsingleinput_{{t_2}_*}) = \kernel_\priorcov(\xsingleinput_{{t_1}_*}, \xsingleinput_{{t_2}_*}) \\ 
        & \quad -\kernel_\priorcov(\xsingleinput_{{t_1}_*}, \xinput) \left( \kernel_\priorcov(\xinput, \xinput) + \noisecov \right)^{-1} \kernel_\priorcov(\xinput, \xsingleinput_{{t_2}_*}).
    \end{split}
\end{align}
Here, $\meanpostpred(\xinput_*)$ represents our model's adapted predictions for the test data, which we can compare to $\youtput_*$ to evaluate the quality of our predictions, for example, via mean squared error (assuming that test data is clean, following \citet{rasmussen}). 
The diagonal of $\covpostpred(\xinput_*, \xinput_*)$ can be interpreted as a per-input level of confidence that captures the ambiguity in making predictions with only a limited amount of context data.

\subsection{The linearization of a neural network yields an expressive linear regression model}
\label{sec:linearization}
As discussed, the choice of feature map $\featuremap$ plays an important role in specifying a linear regression model.
In the deep learning context, recent work has demonstrated that the linear model obtained when linearizing a deep neural network with respect to its weights at initialization, wherein the Jacobian of the network operates as the feature map, can well approximate the behavior of wide nonlinear deep neural networks, especially in the overparameterized regimes \citep{jacot,nonlinear,liu2020linearity,shallow_nns_infinite_width,dnns_infinite_wdith}.
\revise{Furthermore, \citet{maddox} demonstrated that this linearized approximation effectively captures the local geometry of the loss landscape, making it well-suited for uncertainty-aware adaptation.}

Let $\modelsingleinput$ be a neural network $f: \left(\param, \xsingleinput_t \right) \mapsto \ysingleoutput_t$, where $\param \in \R^{\pdim}$ are the parameters of the model, $\xsingleinput \in \R^{\revise{\xdim}}$ is an input and $\ysingleoutput \in \R^{\revise{\ydim}}$ an output.
The linearized network (w.r.t. the parameters) around $\paramlin$ is
\begin{displaymath}
    \modelsingleinput(\param, \xsingleinput_t) - \modelsingleinput(\paramlin, \xsingleinput_t) \approx \jac_\param( \modelsingleinput )(\paramlin, \xsingleinput_t) (\param - \paramlin),
\end{displaymath}
where $\jac_\param(\modelsingleinput)(\cdot, \cdot): \R^\pdim \times \R^\xdim \rightarrow \R^{\ydim \times \pdim}$ is the Jacobian of the network (w.r.t. the parameters).

In the case where the model accepts a batch of $\batchsize$ inputs $\xinput = (\xsingleinput_1, \dots, \xsingleinput_\batchsize)$ and returns $\youtput = (\ysingleoutput_1, \dots, \ysingleoutput_\batchsize)$, we generalize $\modelsingleinput$ to $\model: \R^\pdim \times \R^{\xdim \times \batchsize}  \rightarrow \R^{\ydim \times \batchsize}$, with $\youtput = \model(\param, \xinput)$.
Consequently, we generalize the linearization:
\begin{displaymath}
\model(\param, \xinput) - \model(\paramlin, \xinput) \approx \jac(\paramlin, \xinput) (\param - \paramlin),
\end{displaymath}
where $\jac(\cdot, \cdot): \R^\pdim \times \R^{\xdim \times \batchsize} \rightarrow \R^{\ydim \batchsize \times \pdim}$ is a shorthand for $\jac_\param(g)(\cdot, \cdot)$.
Note that we have implicitly vectorized the outputs, and throughout the work, we will interchange the matrices $\R^{\ydim \times \batchsize}$ and the vectorized matrices $\R^{\ydim \batchsize}$.

This linearization can be viewed as the $\ydim \batchsize$-dimensional linear regression
\begin{equation}
    \label{eq:linearized_network}
    \vz = \featuremap_{\paramlin}(\xinput)^\top \paramcorr \in \R^{\ydim \batchsize},
\end{equation}
where the feature map $\featuremap_{\paramlin}(\cdot): \R^{\xdim \times \batchsize} \rightarrow \R^{\pdim \times \ydim \batchsize}$ is the transposed Jacobian $\jac(\paramlin, \cdot)^\top$.
The parameters of this linear regression $\paramcorr = \left( \param - \paramlin \right)$ are the \textit{correction} to the parameters chosen as the linearization point.
Equivalently, this can be seen as a kernel regression with the kernel $ \kernel_{\paramlin}(\xinput_1,\xinput_2) = \jac(\paramlin, \xinput_1) \jac(\paramlin, \xinput_2)^\top$, which is commonly referred to as the Neural Tangent Kernel (NTK) of the network.
Note that the NTK depends on the linearization point $\paramlin$. 
Building on these ideas, \citet{maddox} show that the NTK obtained via linearizing a DNN \textit{after} it has been trained on a task yields a GP that is well-suited for adaptation and fine-tuning to new, similar tasks. Furthermore, they show that networks trained on similar tasks tend to have similar Jacobians, suggesting that neural network linearization can yield an effective model for multi-task contexts such as meta-learning. In this work, we leverage these insights to construct our parametric functional distribution $\distparam$ via linearizing a neural network model.

\section{Methods}
\label{sec:approach}
In this section, we describe our meta-learning regression algorithm \approach{} and the construction of a parametric functional distribution $\distparam$ that can model the true underlying distribution over tasks $\dist$.
First, we focus on the single cluster case, where a Gaussian process structure on $\distparam$ can effectively model the true distribution of tasks, and detail how we can leverage meta-training data $\dataset$ from a single cluster of tasks to train the parameters $\xi$ of our model.
Next, we will generalize our approach to the multimodal setting, with more than one cluster of tasks. Here, we construct $\distparam$ as a mixture of GPs and develop a training approach that can automatically identify the clusters present in the training dataset without requiring the meta-training dataset to contain any additional structure such as cluster labels.

\subsection{Tractable prior predictive distribution over functions} \label{sec:posterior}
In our approach, we choose $\distparam$ to be the 
GP distribution over functions that arises from a Gaussian prior on the weights of the linearization of a neural network (\eqref{eq:linearized_network}). Consider a particular task $\task$ and a batch of $\batchsize$ context data $(\xcontextinput, \ycontextoutput)$.
The resulting prior predictive distribution, derived from \eqref{eq:prior_pred_dist} after evaluating on the context inputs, is $\youtput | \xcontextinput \sim \normal( \postpredmean, \postpredcov)$, where
\begin{align}
    \label{eq:prior_pred_dist_ntk}
    \postpredmean &= \jac(\paramlin, \xcontextinput) \priormean , \nonumber \\
    \postpredcov &= \jac(\paramlin, \xcontextinput) \priorcov \jac(\paramlin, \xcontextinput)^\top + \noisecov.
\end{align}
In this setup, the parameters $\xi$ of $\distparam$ that we wish to optimize are the linearization point $\paramlin$, and the parameters of the prior over the weights $(\priormean, \priorcov)$.
Given this Gaussian prior, it is straightforward to compute the joint NLL of the context labels $\ycontextoutput$,
\begin{align}
\label{eq:single-nll}
    \mathrm{NLL}(\xcontextinput, \ycontextoutput) & = \frac12 \Big( \left\| \ycontextoutput - \postpredmean \right\|^2_{\postpredcov^{-1}} \nonumber \\
    & \quad + \log\det \postpredcov + \ydim \batchsize \log 2 \pi \Big).
\end{align}
The NLL (a) serves as a loss function quantifying the quality of $\xi$ during training and (b) serves as an uncertainty signal at test time to evaluate whether context data $(\xcontextinput, \ycontextoutput)$ is OoD.
Given this model, \textit{adaptation} is tractable as we can condition this GP on the context data analytically. In addition, we can efficiently make probabilistic predictions by evaluating the mean and covariance of the resulting posterior predictive distribution on the query inputs, using \eqref{eq:post_pred_dist}.

\subsection{Efficient parameterization of the prior covariance}
\label{sec:prior_covariance}
When working with deep neural networks, the number of weights $\pdim$ can easily surpass a million. While it remains tractable to deal with $\paramlin$ and $\priormean$, whose memory footprint grows linearly with $\pdim$, it can quickly become intractable to make computations with (let alone store) a dense prior covariance matrix over the weights $\priorcov \in \R^{\pdim \times \pdim}$. Thus, we must impose some structural assumptions on the prior covariance to scale to deep neural network models.

\textbf{Imposing a unit covariance.}~\, One simple way to tackle this issue would be to remove $\priorcov$ from the learnable parameters $\xi$, i.e., fixing it to be the identity $\priorcov = \mI_{\pdim}$. In this case, $\xi = (\paramlin, \priormean)$. 
This computational benefit comes at the cost of model expressivity, as we lose a degree of freedom in how we can optimize our learned prior distribution $\distparam$. In particular, we are unable to choose a prior over the weights of our model that captures correlations between elements of the feature map.

\textbf{Learning a low-dimensional representation of the covariance.}~\,
An alternative is to learn a low-rank representation of $\priorcov$, allowing for a learnable weight-space prior covariance that can encode correlations. Specifically, we consider a covariance of the form $\priorcov = \proj^\top \diag{\scaling^2} \proj$, where $\proj$ is a fixed projection matrix on an $\sdim$-dimensional subspace of $\R^{\pdim}$, while $\scaling^2$ is learnable.
In this case, the parameters that are learned are $\xi = (\paramlin, \priormean, \scaling)$.
We define $\Scaling := \diag{\scaling^2}$.
The computation of the covariance of the prior predictive (\eqref{eq:prior_pred_dist_ntk}) could then be broken down into two steps:
\begin{displaymath}
\left\{
    \begin{array}{l}
        A := \jac(\paramlin, \xcontextinput) \proj^\top \\
        \jac(\paramlin, \xcontextinput) \priorcov \jac(\paramlin, \xcontextinput)^\top = A \Scaling A^\top 
    \end{array}
\right.
\end{displaymath}
which requires a memory footprint of $O(\pdim(\sdim + \ydim \batchsize) )$, if we include the storage of the Jacobian.
Because $\ydim \batchsize \ll \pdim$ in typical deep learning contexts, it suffices that $\sdim \ll \pdim$ so that it becomes tractable to deal with this new representation of the covariance.

\textbf{A trade-off between feature-map expressiveness and learning a rich prior over the weights.} Note that even if a low-dimensional representation of $\priorcov$ enriches the prior distribution over the weights, it also restrains the expressiveness of the feature map in the kernel by projecting the $\pdim$-dimensional features $\jac(\paramlin, \xinput)$ on a subspace of size $\sdim \ll \pdim$ via $\proj$.
This presents a trade-off: we can use the full feature map, but limit the weight-space prior covariance to be the identity matrix by keeping $\priorcov = \mI$: \trainingid{}. Alternatively, we could learn a low-rank representation of $\priorcov$ by randomly choosing $\sdim$ orthogonal directions in $\R^{\pdim}$, with the risk that they could limit the expressiveness of the feature map if the directions are not relevant to the problem that is considered: \trainingrandom{}.
\yj{To mitigate the issue of selecting random directions,}
we can choose the projection matrix more intelligently and project to the most impactful subspace of the full feature map---in this way, we can reap the benefits of a tunable prior covariance while minimizing the useful features that the projection drops. To select this subspace, we construct this projection map by choosing the top $\sdim$ eigenvectors of the Fisher information matrix (FIM) evaluated on the training dataset $\dataset$: \trainingfim{}. 
\yj{The proposed FIM approach is inspired by \citep{scod},} which demonstrates that the FIM for deep neural networks exhibits rapid spectral decay, suggesting that retaining only a few top eigenvectors suffices to encode an expressive task-tailored prior.
\yj{In the following sections, we use \approach{} to refer to \trainingfim{}.
The pseudo-code for \trainingfim{} is described in Algorithm~\ref{alg:meta_training_learnt_cov}.
See Appendix~\ref{app:fim} for more details, including the pseudo-codes for \trainingid{} and \trainingrandom{}.}
We also provide a detailed computational complexity analysis in the Appendix~\ref{sec:appendix_complexity}, which shows that our method's \textbf{cost scales linearly} with the number of model parameters ($P$), comparable to MAML, ensuring its practicality for common meta-learning applications.

\begin{algorithm}[t]
\caption{\footnotesize \approach{}-F}
\footnotesize
\label{alg:meta_training_learnt_cov}
\begin{algorithmic}[1]
        \State Find intermediate $\paramlin$, $\priormean$ with \trainingid{} \Comment{see Alg.~\ref{alg:meta_training_identity}}
        \State Find $\proj$ via \Call{FIMProj}{\sdim}; initialize $\scaling$. \Comment{see Alg.~\ref{alg:fim_proj}}
        \ForAll{epoch}
            \State Sample $\ntasksperepoch$ tasks $\{ \task, (\xcontextinput, \ycontextoutput) \}_{i=1}^{i=\ntasksperepoch}$
            \ForAll{$\task, (\xcontextinput, \ycontextoutput)$}
                \State $\Sigma_i \gets \jac \proj^\top \diag{\scaling^2} \proj \jac^\top + \noisecov$ \Comment{{\tiny $\jac = \jac(\paramlin, \xcontextinput)$}}
\State $NLL_i \gets \Call{GaussNLL}{\ycontextoutput;\, \jac\priormean,\, \Sigma_i}$
            \EndFor
            \State Update $\paramlin$, $\priormean$, $\scaling$ with $\nabla_{\paramlin \cup \priormean \cup \scaling} \sum_i NLL_i$
        \EndFor
\end{algorithmic}
\end{algorithm}

\subsection{Generalization to a mixture of Gaussians}
\label{sec:mixture}

When learning on multiple clusters of tasks, $\dist$ can become non-unimodal, and thus cannot be accurately described by a single GP.
Instead, we can capture this multimodality by structuring $\distparam$ as a \textit{mixture} of Gaussian processes.

\textbf{Building a more general structure.}~\, We assume that at train time, a task $\task$ comes from any cluster $\left\{\cluster_j \right\}_{j=1}^{j=\nclusters}$ with equal probability.
Thus, we choose to construct $\distparam$ as an equal-weighted mixture of $\nclusters$ Gaussian processes.

For each element of the mixture, the structure is similar to the single cluster case, where the parameters of the cluster's weight-space prior are given by $(\priormean_j, \priorcov_j)$. We choose to have both the projection matrix $\proj$ and the linearization point $\paramlin$ (and hence, the feature map $\featuremap(\cdot) = \jac(\paramlin,\cdot)$) shared across the clusters. This yields improved computational efficiency, as we can compute the projected features once, simultaneously, for all clusters.
This yields the parameters $\xi_\nclusters = (\paramlin, \proj, (\priormean_1, \scaling_1), \ldots, (\priormean_\nclusters, \scaling_\nclusters))$.

This can be viewed as a mixture of linear regression models, with a common feature map but 
separate, independent prior distributions over the weights for each cluster. These separate distributions are encoded using the low-dimensional representations $\Scaling_j$ for each $\priorcov_j$. 
Notice how this is a generalization of the single cluster case, for when $\nclusters=1$, $\distparam$ becomes a Gaussian and $\xi_\nclusters = \xi$\footnote{In theory, it is possible to drop $\proj$ and extend the identity covariance case to the multi-cluster setting; however, this leads to each cluster having an identical covariance function, and thus is not effective at modeling heterogeneous behaviors among clusters.}.

\textbf{Prediction and likelihood computation.}~\, The NLL of a batch of inputs under this mixture model can be computed as
\begin{align}
    \label{eq:nll_mixture}
    & \mathrm{NLL}_{\text{mixt}}(\xcontextinput, \ycontextoutput) = \log \alpha \nonumber \\
    & - \mathrm{LSE} (-\mathrm{NLL}_1(\xcontextinput, \ycontextoutput), \ldots, -\mathrm{NLL}_\alpha(\xcontextinput, \ycontextoutput)),
\end{align}
\revise{where $\mathrm{NLL}_j(\xcontextinput, \ycontextoutput)$ is the NLL with respect to each individual Gaussian, as computed in \eqref{eq:single-nll}, and  $\mathrm{LSE}(\cdot) := \log\add\exp(\cdot)$ computes the logarithm of the sum of the exponential of these arguments, avoiding underflow issues.}

To make exact predictions, we would require conditioning this mixture model. To simplify this, we propose to first \textit{infer the cluster} from which a task comes from, by identifying the Gaussian $\gaussian_{j_0}$ that yields the highest likelihood for the context data $\left( \xcontextinput, \ycontextoutput \right)$. Then, we can \textit{adapt} by conditioning $\gaussian_{j_0}$ with the context data and finally \textit{infer} by evaluating the resulting posterior distribution on the queried inputs $\xqueryinput$.

\subsection{Meta-training the Parametric Task Distribution}
The key to our meta-learning approach is to estimate the quality of $\distparam$ via the NLL of context data from training tasks, and use its gradients to update the parameters of the distribution $\xi$.
Optimizing this loss over tasks in the dataset draws $\distparam$ closer to the empirical distribution present in the dataset, and hence towards the true distribution $\dist$.

\textbf{Computing the likelihood.}~\,
In the algorithms, the function \Call{GaussNLL}{$\ycontextoutput$; $m$, $K$} stands for NLL of $\ycontextoutput$ under the Gaussian $\normal(m, K)$ (see \eqref{eq:single-nll}).
In the mixture case, we instead use \Call{MixtNLL}{}, which wraps \eqref{eq:nll_mixture} and calls \Call{GaussNLL}{} for the individual NLL computations.
In this case, $\priormean$ becomes $\{\priormean_j\}_{j=1}^{j=\nclusters}$ and $\scaling$ becomes $\{\scaling_j\}_{j=1}^{j=\nclusters}$ when applicable.

\textbf{Finding the FIM-based projections.}~\,
The FIM-based projection matrix aims to identify the elements of $\featuremap = \jac(\paramlin, \xinput)$ that are most relevant for the problem.
However, this feature map evolves during training, because it is $\paramlin$-dependent.
How do we ensure that the directions we choose for $\proj$ remain relevant during training?
We leverage results from \citet{ntk_evolution}, stating that the NTK (the kernel associated with the Jacobian feature map) changes significantly at the beginning of training and that its evolution slows down as training goes on. This suggests that as a heuristic, we can compute the FIM-based directions after partial training, as they are unlikely to deviate much after the initial training. 
For this reason, \trainingfim{}
(Algorithm~\ref{alg:meta_training_learnt_cov}) first calls \trainingid{} (Algorithm~\ref{alg:meta_training_identity}) before computing the FIM-based $\proj$ that yields intermediate parameters $\paramlin$ and $\priormean$.
Then the usual training takes place with the learning of $\scaling$ in addition to $\paramlin$ and $\priormean$.

\begin{figure}[t]
     \centering
     \begin{subfigure}[t]{0.42\textwidth}
         \centering
         \includegraphics[width=\textwidth]{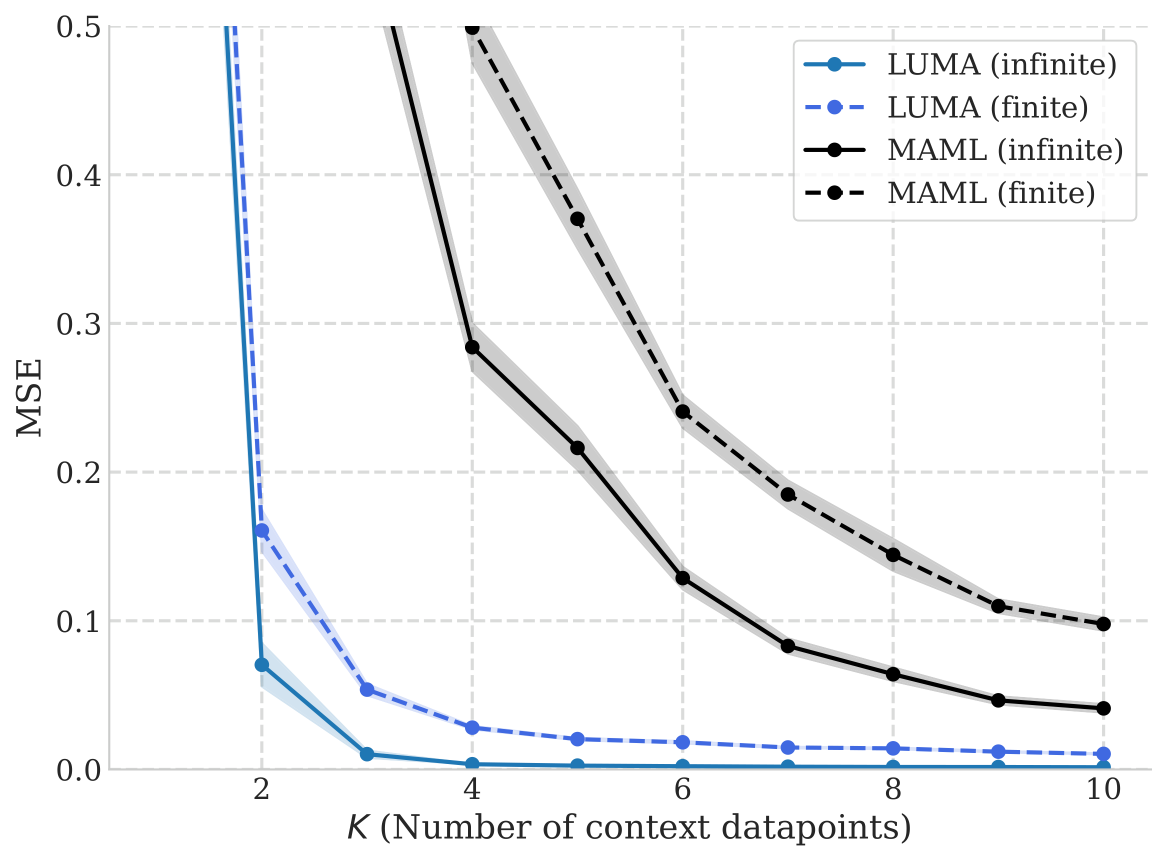}
         \caption{Unimodal}
         \label{fig:single-mse}
     \end{subfigure}
     \begin{subfigure}[t]{0.42\textwidth}
         \centering
         \includegraphics[width=\textwidth]{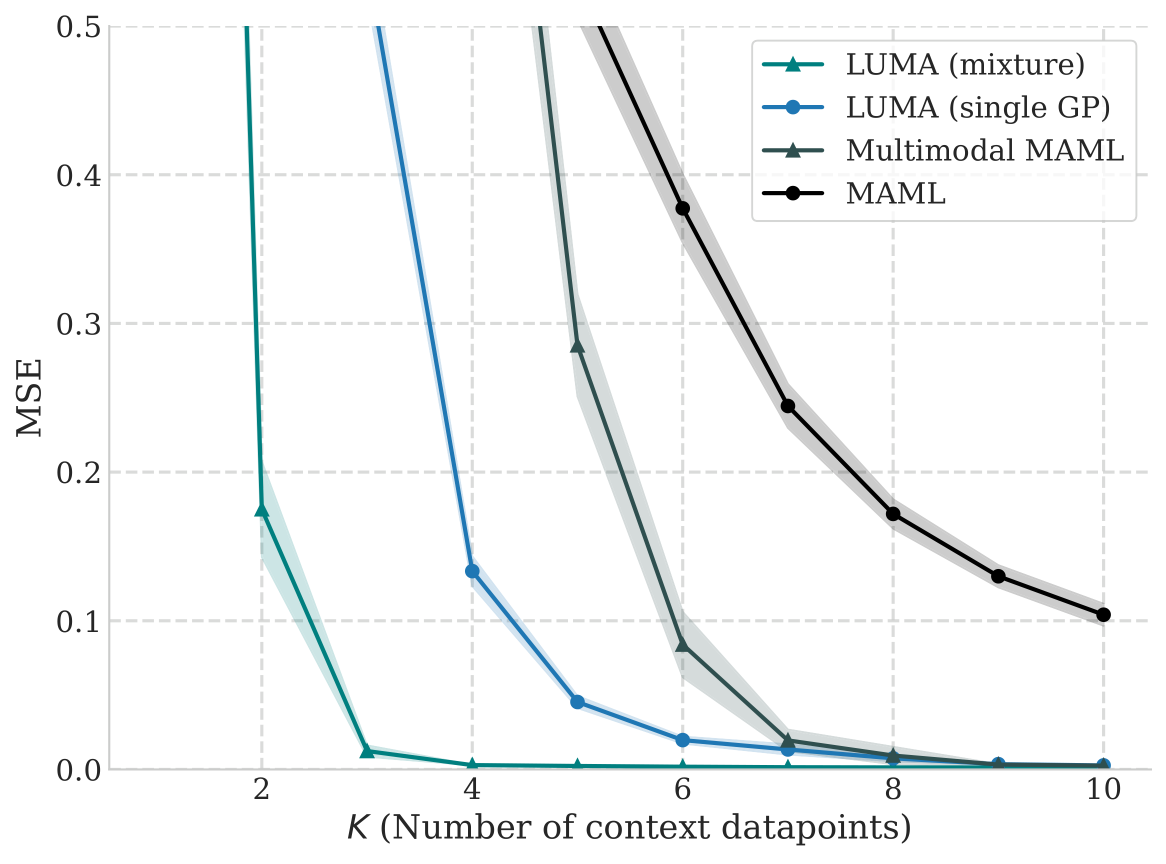}
         \caption{Multimodal}
         \label{fig:multi-mse}
     \end{subfigure}
    \caption{
    MSE on predictions: We evaluate the performance of \approach{} with varying numbers of context datapoints, $\batchsize$.
    In the unimodal setting, \approach{} trained on a finite task dataset outperforms the baseline, even when the baseline is trained on an infinite task dataset.
    In the multimodal case, not only does our multimodal \approach{} (mixture) outperform the baselines, but even the single-GP \approach{} still surpasses them as well.
    }
    \label{fig:mse_comparison}
\end{figure}

\yj{
\section{Numerical Analysis}
\label{sec:results}

In this section, we evaluate the empirical efficacy of the proposed framework by examining the following key aspects of \approach:
(1) Comparative accuracy of the proposed probabilistic framework against baselines in both unimodal and multimodal settings, (2) OoD detection performance of the proposed method, and (3) ablation study regarding the trade-off between learning a rich prior distribution and maintaining a complete feature map.

\paragraph{Comparison with baselines.}
To show the efficacy of \approach{}, we compare it to the closely related prominent meta-learning frameworks, i.e., MAML \citep{maml} and Multimodal MAML (MMAML) \citep{mmaml}\revise{, as well as additional baselines, namely, Conditional Neural Processes (CNP) \citep{cnp}, Transformer Neural Processes (TNP-D) \citep{tnpd}, and Deep Kernel Transfer (DKT) \citep{dkt}}.
First, we consider a unimodal setting with the task distribution consisting of sinusoids of varying amplitude and phase.
We use a neural network with 2 hidden layers, 40 neurons each, with a ReLU non-linearity, and a single GP structure of $\distparam$ for \approach{}.
In a unimodal setting, we also compare the results between training with an infinite amount of available sine tasks (infinite task dataset), and with a finite amount of available tasks (finite task dataset).
We then extend the empirical analysis to a multimodal setting with training data consisting of sinusoids as well as lines with varying slopes.
Details on the problems can be found in Appendix~\ref{app:train-details-single}, and the training and test details for unimodal and multimodal cases are in Appendix~\ref{app:problem-details} and Appendix~\ref{app:train-details-multi}, respectively.

As shown in Figure~\ref{fig:single-mse}, \emph{the empirical evidence confirms that \approach{} outperforms MAML, particularly in the low-data regime}; it achieves much better generalization particularly when we have a small number of context samples, $\batchsize$.
Moreover, \approach{} trained with a finite task dataset performs comparably to the one with an infinite dataset and predicts better than the baselines. This shows the robustness of \approach{}, capturing the common features of the tasks even with limited meta-training.
Examples of test predictions are in Figure~\ref{fig:single-predictions-full} in the Appendix.

The same trend is further highlighted in the multimodal setting. Notably, \approach{} with a single GP structure outperforms both MAML and MMAML in prediction. This highlights the strength of our probabilistic approach for multimodal meta-learning: even when the probabilistic assumptions are not perfectly aligned, the predictions remain accurate and surpass baseline methods. Moreover, this performance gap widens when incorporating a mixture structure into our framework.

Finally, we conducted a comparative analysis against probabilistic meta-learning baselines, PLATIPUS and BaMAML\revise{, as well as CNP, TNP-D, and DKT}. As shown in Table~\ref{tab:mse_comparison_k10}, the results underscore the superior performance of our proposed method, which consistently achieves a lower prediction error across all evaluated tasks and datasets. \revise{In the unimodal infinite-task setting, neural processes (CNP, TNP-D) perform competitively, but their performance degrades substantially in the finite-task setting ($\ntraintasks=10$), whereas \approach{} maintains high accuracy, demonstrating superior sample efficiency across tasks. In the multimodal setting, our mixture model achieves the lowest error by explicitly capturing the task structure (see Table~\ref{tab:mse_comparison_multimodal} in the Appendix for full results).} We assess the reliability of their uncertainty estimates in the subsequent analysis section.

\paragraph{Vision Regression Tasks.} \label{sec:deep}

In addition to our analysis with a shallow MLP, we extend our study to a more complex scenario using a deeper network for a unimodal vision meta-learning problem, ShapeNet1D \citep{what-matters}, which aims to predict object orientations from the image.
In this problem, each task consists of context data comprising images of the same object captured at different orientations, while the query inputs are additional images of the same object with unknown orientations to be predicted.
Details on the problems and the datasets can be found in Appendix.
As shown in Table~\ref{tab:mse_comparison_k10}, \approach{} generalizes well to deeper networks. \revise{DKT excels at larger context sizes ($\batchsize \geq 10$) by leveraging its CNN backbone, while \approach{} shows relatively more stable degradation in the extreme low-data regime ($\batchsize=5$).}

\begin{table}[htb!]
\centering
\setlength{\tabcolsep}{1mm}
\footnotesize
\resizebox{\columnwidth}{!}{%
\revise{
\begin{tabular}{l|ccc}
\toprule
~ & \textbf{Unimodal} & \textbf{Multimodal} & \textbf{Vision} \\
\midrule
MAML & 0.2172 / 0.0314 & 0.5324 / 0.1041 & 22.41 / 17.42 \\
PLATIPUS & 0.2466 / 0.0680 & 0.4979 / 0.1160 & 57.74 / 55.08 \\
BaMAML & 0.4333 / 0.1359 & 0.9964 / 0.3292 & 21.80 / 17.64 \\
CNP & 0.0485 / 0.0189 & 0.1402 / 0.0311 & 22.31 / 19.70 \\
TNP-D & 0.1324 / 0.0186 & 0.1253 / 0.0196 & 89.75 / 88.74 \\
DKT & 3.2730 / 0.2122 & 2.4779 / 0.1760 & 21.32 / \textbf{3.920} \\
\textbf{Ours} & \textbf{0.0026} / \textbf{0.0015} & \textbf{0.0024} / \textbf{0.0012} & \textbf{18.94} / 7.684 \\

\bottomrule
\end{tabular}
}
}
\caption{Prediction error comparison (MSE for regression; angular error for vision tasks) with $\batchsize$=5/10. Our method robustly outperforms the baselines on regression tasks, especially in low-data regimes (e.g., $\batchsize$=5).}
\label{tab:mse_comparison_k10}
\end{table}

\paragraph{OoD detection performance.}

To further examine the effectiveness of \approach{}, we evaluate its OoD detection performance.
Remark that our framework provides an analytical posterior distribution, allowing us to compute the AUC-ROC score using the NLL of the context inputs with respect to $\distparam$.
We compare its reliability to the probabilistic baselines that work on sampling-based uncertainty estimation\revise{, as well as to CNP, TNP-D, and DKT}.
For this analysis, we consider a cluster of sine tasks, one of the linear tasks, and one of the quadratic tasks. %
In the unimodal setting, sine tasks are in-distribution (ID), while others are OoD. In the multimodal setting, sine and linear tasks are ID, with the quadratic task as OoD.

As illustrated in Table~\ref{tab:auc_comparison_k10}, across the different settings, the proposed framework achieves nearly perfect OoD detection accuracy even with a limited number of context data points (e.g., $K$=5).
This further demonstrates the efficacy of our mixture-of-GPs structure in a multimodal setting. %
\revise{In the unimodal setting, CNP and TNP-D achieve comparable near-perfect detection. However, in the multimodal setting, their performance drops to near-random (AUC $\approx 0.5$), as their encoder maps context sets to a single latent representation, making it difficult to distinguish OoD tasks from in-distribution tasks of a different mode. DKT performs poorly across all settings, suggesting that its input-level uncertainty is not well-suited for task-level OoD detection. In contrast, our mixture-of-GPs formulation with tractable per-component likelihoods achieves near-perfect detection (AUC $>0.99$) even in the multimodal case.}

\begin{table}[h]
\centering
\setlength{\tabcolsep}{1mm}
\small
\revise{
\begin{tabular}{l|cc}
\toprule
\textbf{Method} & \textbf{Unimodal} & \textbf{Multimodal} \\
\midrule
PLATIPUS & 0.8199 / 0.9438 & 0.6680 / 0.8363 \\
BaMAML & 0.6705 / 0.8474 & 0.5279 / 0.5786 \\
CNP & 0.9990 / \textbf{1.0000} & 0.5090 / 0.5110 \\
TNP-D & 0.9960 / \textbf{1.0000} & 0.5220 / 0.5340 \\
DKT & 0.2530 / 0.4710 & 0.1950 / 0.4230 \\
\textbf{Ours} & \textbf{1.0000} / \textbf{1.0000} & \textbf{0.9940} / \textbf{1.0000} \\
\bottomrule
\end{tabular}
}
\caption{Out-of-distribution detection performance comparison. AUC-ROC scores with $\batchsize$=5/10 are reported. \approach{} achieves near-perfect scores across all settings. \revise{CNP and TNP-D perform well in the unimodal case but degrade to near-random in the multimodal setting.}}
\label{tab:auc_comparison_k10}
\end{table}

\paragraph{Trade-off analysis.}
We provide an in-depth analysis comparing the performance between \trainingid{}, \trainingrandom{}, and \trainingfim{} to study a trade-off between learning a rich prior distribution over the weights and maintaining the full expressivity of the network.
As shown in Figures~\ref{fig:single-mse-all} and \ref{fig:single-auc-all}, \trainingrandom{} and \trainingfim{} outperform \trainingid{}, reflecting a higher-quality learned prior (see Appendix for details).
For shallow networks, having a rich prior over the weights could be more beneficial than maintaining the full expressiveness of the feature map, making both \trainingrandom{} and \trainingfim{} preferable. 
However, in deep networks (Figure~\ref{fig:vision-mse-all}), preserving an expressive feature map becomes crucial, favoring \trainingid{} and \trainingfim{}. Overall, the FIM-based approach, \trainingfim{}, is the most robust, consistently achieving superior performance.
}
\begin{figure}[t!]
     \centering
     \begin{subfigure}[t]{0.41\textwidth}
         \centering
         \includegraphics[width=\textwidth]{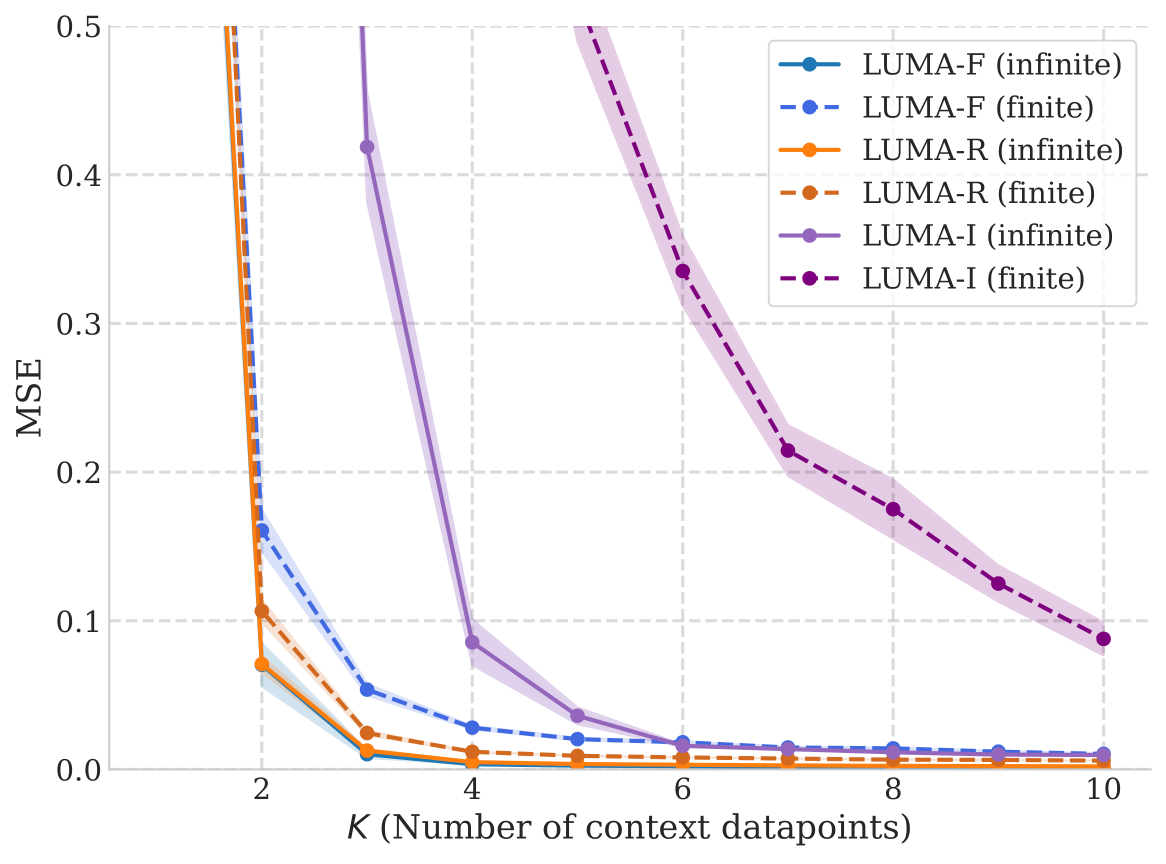}
         \caption{Sinusoid Task}
         \label{fig:single-mse-all}
     \end{subfigure}
     \begin{subfigure}[t]{0.41\textwidth}
         \centering
         \includegraphics[width=\textwidth]{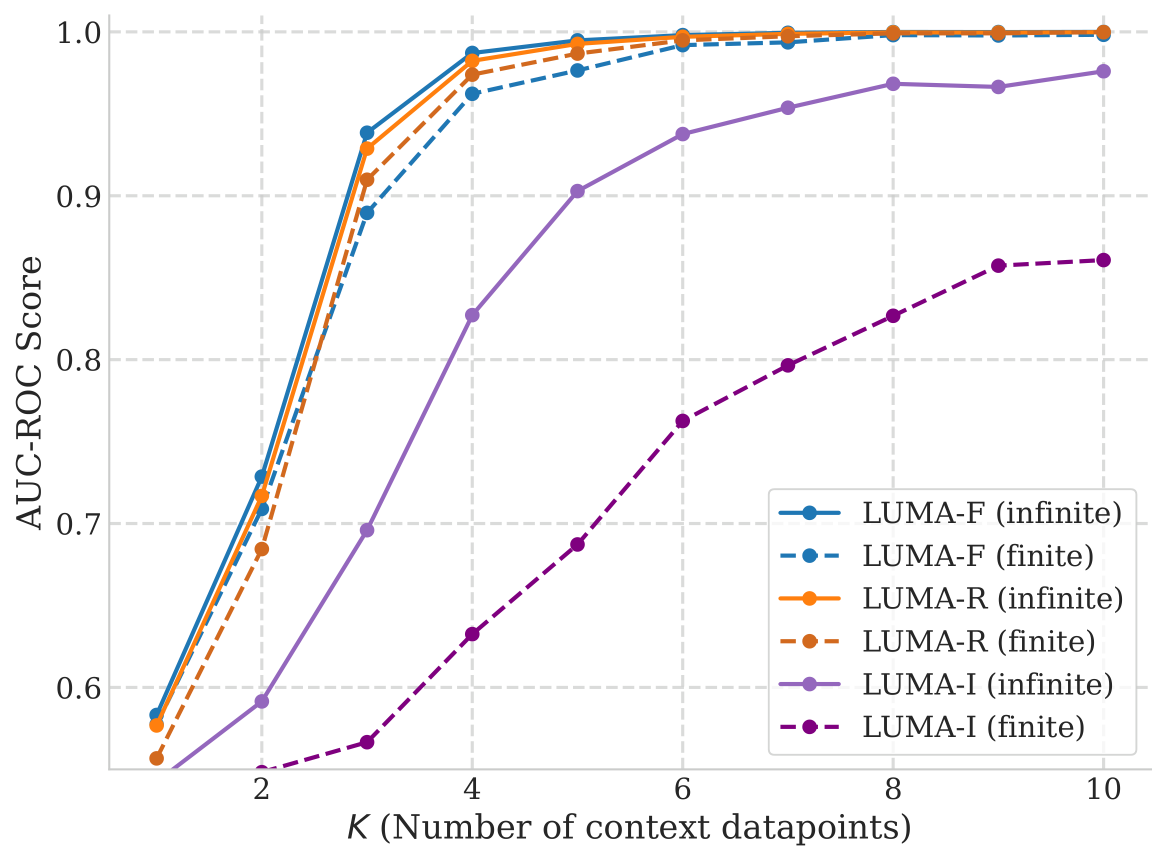}
         \caption{OoD Detection}
         \label{fig:single-auc-all}
     \end{subfigure}
    \label{fig:trade-off}
     \begin{subfigure}[t]{0.41\textwidth}
         \centering
         \includegraphics[width=\textwidth]{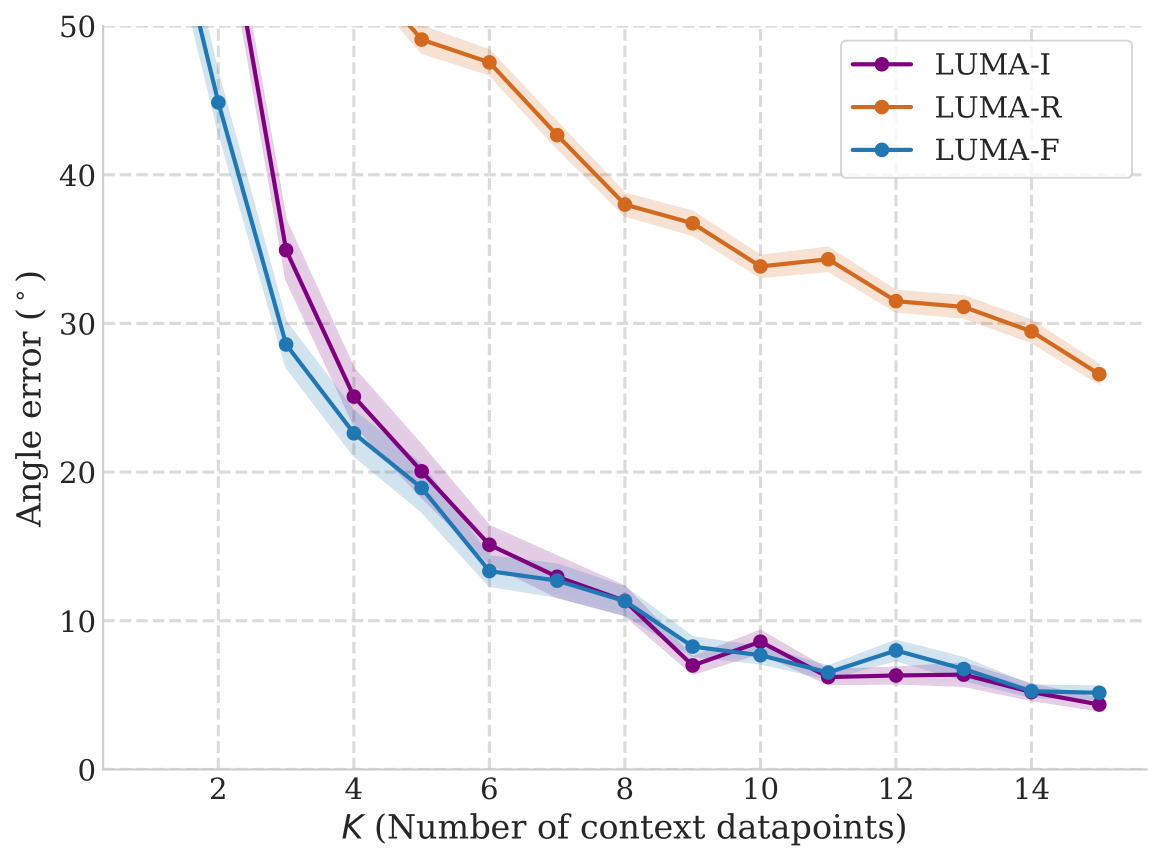}
         \caption{Vision Task}
         \label{fig:vision-mse-all}
     \end{subfigure}
    \caption{Trade-off analysis with \approach{} variants.}
\end{figure}

\revise{\paragraph{Discussion.} Our experiments reveal that different meta-learning paradigms exhibit complementary strengths. Neural processes (CNP, TNP-D) are competitive in the infinite-task setting but degrade substantially when meta-training tasks are limited. DKT leverages its deep feature extractor effectively on vision tasks with sufficient context. In contrast, \approach{} demonstrates the most robust performance in low-data regimes and with limited meta-training data points.
For OoD detection, our advantage is most pronounced in the multimodal setting, where all other baselines fail to reliably distinguish OoD tasks from in-distribution tasks of a different mode, thanks to our mixture-of-GPs formulation with tractable per-component likelihoods.
These findings suggest that \approach{} is particularly well suited to safety-critical few-shot regression where reliable uncertainty estimates and OoD detection are essential, such as autonomous vehicle dynamics adaptation and biomedical forecasting.}

\section{Conclusion}\label{sec:conclusion}

We proposed \approach{}, a meta-learning algorithm that models the underlying task distribution using a parametric and tunable distribution, leveraging Bayesian inference with linearized neural networks.
By incorporating a Fisher-information-based parameterization, \approach{} strikes an effective balance between scalability and expressivity.
We demonstrated that (1) our approach makes efficient probabilistic predictions on in-distribution tasks, (2) it is capable of effectively detecting OoD context data, and that (3) both of these findings continue to hold in the multimodal task-distribution setting.

\paragraph{Future work.}
There are several interesting avenues for future work. Our current framework, built on a Gaussian likelihood, is tailored for regression tasks; generalizing our approach to non-Gaussian likelihoods would enable \approach{} to be used for classification, a direction we believe is highly promising. Furthermore, while our experiments on synthetic and vision-based tasks demonstrate the core strengths of our method, future work could involve evaluation on a broader range of large-scale meta-learning benchmarks to further validate its effectiveness and scalability.

We hope this work encourages the community to further explore the intersection of linearized neural networks and probabilistic meta-learning. We believe the principles of tractable Bayesian inference presented here can serve as a strong foundation for developing more robust, uncertainty-aware, and general-purpose learning systems.

\section*{Acknowledgments}
\revise{The authors acknowledge the MIT SuperCloud and Lincoln Laboratory Supercomputing Center for providing computing resources that have contributed to the results reported within this paper. This work was supported in part by the MIT-IBM Watson AI Lab, the MIT-Google Program for Computing Innovation, the MIT-Amazon Science Hub, Netflix, and Jane Street.}

\bibliography{bibliography}

\section*{Checklist}

The checklist follows the references. For each question, choose your answer from the three possible options: Yes, No, Not Applicable.  You are encouraged to include a justification to your answer, either by referencing the appropriate section of your paper or providing a brief inline description (1-2 sentences). 
Please do not modify the questions.  Note that the Checklist section does not count towards the page limit. Not including the checklist in the first submission won't result in desk rejection, although in such case we will ask you to upload it during the author response period and include it in camera ready (if accepted).

\begin{enumerate}

  \item For all models and algorithms presented, check if you include:
  \begin{enumerate}
    \item A clear description of the mathematical setting, assumptions, algorithm, and/or model. Yes
    \item An analysis of the properties and complexity (time, space, sample size) of any algorithm. Yes
    \item (Optional) Anonymized source code, with specification of all dependencies, including external libraries. Yes
  \end{enumerate}

  \item For any theoretical claim, check if you include:
  \begin{enumerate}
    \item Statements of the full set of assumptions of all theoretical results. Yes
    \item Complete proofs of all theoretical results. Yes
    \item Clear explanations of any assumptions. Yes
  \end{enumerate}

  \item For all figures and tables that present empirical results, check if you include:
  \begin{enumerate}
    \item The code, data, and instructions needed to reproduce the main experimental results (either in the supplemental material or as a URL). Yes
    \item All the training details (e.g., data splits, hyperparameters, how they were chosen). Yes
    \item A clear definition of the specific measure or statistics and error bars (e.g., with respect to the random seed after running experiments multiple times). Yes
    \item A description of the computing infrastructure used. (e.g., type of GPUs, internal cluster, or cloud provider). Yes
  \end{enumerate}

  \item If you are using existing assets (e.g., code, data, models) or curating/releasing new assets, check if you include:
  \begin{enumerate}
    \item Citations of the creator If your work uses existing assets. Yes
    \item The license information of the assets, if applicable. Yes
    \item New assets either in the supplemental material or as a URL, if applicable. Yes
    \item Information about consent from data providers/curators. Not Applicable
    \item Discussion of sensible content if applicable, e.g., personally identifiable information or offensive content. Not Applicable
  \end{enumerate}

  \item If you used crowdsourcing or conducted research with human subjects, check if you include:
  \begin{enumerate}
    \item The full text of instructions given to participants and screenshots. Not Applicable
    \item Descriptions of potential participant risks, with links to Institutional Review Board (IRB) approvals if applicable. Not Applicable
    \item The estimated hourly wage paid to participants and the total amount spent on participant compensation. Not Applicable
  \end{enumerate}

\end{enumerate}

\newpage
\onecolumn

\title{Tractable Uncertainty-Aware Meta-Learning \\(Supplementary Material)}
\date{}
\maketitle
\thispagestyle{empty}

\appendix

\section{Related Work}
\label{sec:related_work_appendix}
\paragraph{Bayesian inference with linearized DNNs.}
Bayesian inference with neural networks is often intractable because the posterior predictive has rarely a closed-form expression.
Whereas \approach{} linearizes the network to allow for practical Bayesian inference, existing work has used other approximations to tractably express the posterior.
For example, it has been shown that in the infinite-width approximation, the posterior predictive of a Bayesian neural network behaves like a GP \citep{shallow_nns_infinite_width, dnns_infinite_wdith}. This analysis can in some cases yield a good approximation to the Bayesian posterior of a DNN \citep{cnns_infinite_width}.
It is also common to use Laplace's method to approximate the posterior predictive by a Gaussian distribution and allow practical use of the Bayesian framework for neural networks.
This approximation relies in particular on the computation of the Hessian of the network: this is in general intractable, and most approaches use the so-called Gauss-Newton approximation of the Hessian instead \citep{laplace_scalable}.
Recently, it has been shown that Laplace's method using the Gauss-Newton approximation is equivalent to working with a certain linearized version of the network and its resulting posterior GP \citep{laplace_linearization}.

Bayesian inference is applied in a wide range of subjects. For example, recent advances in transfer learning have been possible thanks to Bayesian inference with linearized neural networks.
\citet{maddox} have linearized pre-trained networks and performed domain adaptation by conditioning the prior predictive with data from the new task: the posterior predictive is then used to make predictions. Our approach leverages a similar adaptation method and demonstrates how the prior distribution can be learned in a meta-learning setup.

\paragraph{Meta-learning.}
MAML is a meta-learning algorithm that uses as adaptation a few steps of gradient descent \citep{maml}.
It has the benefit of being model-agnostic (it can be used on any model for which we can compute gradients w.r.t. the weights), whereas \approach{} requires the model to be a differentiable regressor.
MAML has been further generalized to probabilistic meta-learning models such as PLATIPUS or BaMAML \citep{bayesian_maml, probabilistic_maml}, where the simple gradient descent step is augmented to perform approximate Bayesian inference. These approaches, like ours, learn (during meta-training) and make use of (at test-time) a prior distribution over the weights. \textbf{In contrast, however, \approach{} \revise{performs analytically tractable Bayesian inference on a linearized model} at test-time}.
\yj{Therefore, unlike other probabilistic frameworks that estimate the posterior predictive distribution through sampling, our method yields an \emph{analytically tractable} posterior distribution.}
MAML has also been improved for multimodal meta-learning via MMAML \citep{mmaml, revisit_mmaml}. Similarly to our method, they add a step to identify the cluster from which the task comes from \citep{mmaml}.
OoD detection in meta-learning has been studied by \citet{ood_maml}, who built upon MAML to perform OoD detection in the classification setting, to identify unseen classes during training.
\citet{meta_learning_ood} also implemented OoD detection for classification, by learning a Gaussian mixture model on a latent space. 
\approach{} extends these ideas to the regression case, aiming to identify when test data is drawn from an unfamiliar function.

ALPaCA is a Bayesian meta-learning algorithm for neural networks, where only the last layer is Bayesian \citep{alpaca}.
This framework yields an exact linear regression that uses as feature map the activations right before the last layer.
\textbf{Our work is a generalization of ALPaCA}, in the sense that \approach{} restricted to the last layer matches ALPaCA's approach.
The link between these methods is further discussed in Appendix~\ref{app:link_with_alpaca}. More broadly, control-oriented meta-learning approaches have been developed, e.g., by \cite{richards2023control,tang2025meta}, though they are not probabilistic.

\revise{\paragraph{Sparse Gaussian neural processes.} \citet{sgnp} extend the sparse variational GP \citep{titsias2009variational} framework to the meta-learning setting by training an encoder function that predicts task-specific inducing points from context data, enabling rapid sparse GP inference on new tasks without per-task optimization. While both SGNP and our \approach{} yield analytically tractable posterior predictive distributions, the two methods are fundamentally different: SGNP predicts inducing points that summarize the context data, whereas \approach{} directly infers a posterior distribution over functions in the weight space of a linearized neural network. Another notable difference is that \approach{} provides an analytically tractable prior predictive (\eqref{eq:single-nll}), enabling principled task-level OoD detection. In contrast, SGNP does not compute the likelihood of context data; thus, it's not suitable for OoD detection.}

\paragraph{Meta-learning vs.\ fine-tuning approaches}
Foundation models are large-scale pre-trained neural networks trained on vast amounts of unlabeled data from diverse domains \citep{devlin2019bert, brown2020language, bommasani2021opportunities}. These models serve as general-purpose backbones for a wide range of downstream tasks and have significantly influenced research in few-shot learning.

A widely used approach for adapting foundation models is fine-tuning, where the model's parameters are further updated on task-specific data \citep{howard2018universal, houlsby2019parameter, lester2021power}. Although this method is straightforward and often effective, it can be computationally expensive and prone to performance degradation, particularly when only a small number of labeled examples are available. 
Alternatively, in-context learning provides examples of the desired task within the models input prompt, enabling generalization to new queries without updating model parameters \citep{brown2020language}. However, meta-learning offers a more principled framework for adapting to families of related tasks, allowing for rapid generalization even in low-data regimes. Additionally, meta-learning approaches tend to be more robust to domain shifts, as the meta-training phase involves adaptation to diverse tasks and conditions, explicitly preparing the model to generalize effectively.

\begin{algorithm}[t]
\caption{\footnotesize \trainingid: meta-training with identity prior covariance}
\footnotesize
\label{alg:meta_training_identity}
\begin{algorithmic}[1]
        \State Initialize $\paramlin$, $\priormean$.
        \ForAll{epoch}
            \State Sample $\ntasksperepoch$ tasks $\{ \task, (\xcontextinput, \ycontextoutput) \}_{i=1}^{i=\ntasksperepoch}$
            \ForAll{$\task, (\xcontextinput, \ycontextoutput)$}
                \State $NLL_i \gets \Call{GaussNLL}{\ycontextoutput; \jac\priormean,~ \jac\jac^\top + \noisecov}$ \Comment{$\jac = \jac(\paramlin, \xcontextinput)$}
            \EndFor
            \State Update $\paramlin$, $\priormean$ with $\nabla_{\paramlin \cup \priormean} \sum_i NLL_i$
        \EndFor
\end{algorithmic}
\end{algorithm}

\begin{algorithm}[t]
\caption{\footnotesize \trainingrandom{} and \trainingfim{}: meta-training with a learnt covariance}
\footnotesize
\begin{algorithmic}[1]
        \If{using random projections}
            \State Find random projection $\proj$
            \State Initialize $\paramlin$, $\priormean$, $\scaling$
        \ElsIf{using FIM-based projections}
            \State Find intermediate $\paramlin$, $\priormean$ with \trainingid{} \Comment{see Alg.~\ref{alg:meta_training_identity}}
            \State Find $\proj$ via \Call{FIMProj}{\sdim}; initialize $\scaling$. \Comment{see Alg.~\ref{alg:fim_proj}}
        \EndIf
        \ForAll{epoch}
            \State Sample $\ntasksperepoch$ tasks $\{ \task, (\xcontextinput, \ycontextoutput) \}_{i=1}^{i=\ntasksperepoch}$
            \ForAll{$\task, (\xcontextinput, \ycontextoutput)$}
                \State $NLL_i \gets \Call{GaussNLL}{\ycontextoutput; \jac\priormean,~ \jac \proj^\top \diag{\scaling^2} \proj \jac^\top + \noisecov}$ \Comment{$\jac = \jac(\paramlin, \xcontextinput)$}
            \EndFor
            \State Update $\paramlin$, $\priormean$, $\scaling$ with $\nabla_{\paramlin \cup \priormean \cup \scaling} \sum_i NLL_i$
        \EndFor
\end{algorithmic}
\end{algorithm}

\section{A tractable way of finding the perturbation directions in weight space that impact the most the predictions of an entire dataset}
\label{app:fim}

Deep neural networks have a large number of parameters, making the feature map $\featuremap_{\paramlin}$ high-dimensional.
However, recent work has shown that only a small subspace of the weight space is impactful.
For example, to perform continual learning, \citet{ogd} leverage the fact that it is sufficient to update the parameters orthogonally to a few directions only to avoid catastrophic forgetting.
\citet{sagun} have shown that the Hessian of a deep neural network can be summarized in a few number of directions, due to rapid spectral decay.
This encourages finding a method to extract these meaningful directions of the weight space.

\subsection{Link with the Fisher Information Matrix}

We define these main directions as the ones that have the most impact on the predictions of a whole dataset.
To find them, we first find a way to quantify the influence of an infinitesimal weight perturbation.
Using the second-order approximation of that quantity, we then describe in the deep-learning context a tractable way to find the directions.

\paragraph{Setting} We take the same setting as Section Background, and we describe a method to quantify the influence of a parameter perturbation $\parampert$ on the predictions of a dataset of tasks $\dataset$. To do so, we leverage a probabilistic interpretation of the model: we assume a Gaussian pdf over the observations for given inputs and parameters $p_{\param}(\youtput \vert \xinput) \sim \normal(\model(\param, \xinput), ~\noisecov)$, where the covariance of the noise $\noisecov$ is diagonal $\noisecov = \stdnoise^2 \mI$ (just as in Section Background).

\textbf{Perturbation of the prediction of a batch of inputs.} Before quantifying the influence of the perturbation on the predictions of the whole task dataset, we do it for the prediction of a batch of inputs $\model(\paramlin, \xinput)$.

We borrow the method from \citet{scod}: we quantify the influence of a parameter perturbation by computing the Kullback-Leibler divergence between $p_{\paramlin}(\youtput \vert \xinput)$ and $p_{\paramlin + \parampert}(\youtput \vert \xinput)$. The expansion is:
\begin{equation}
    \label{eq:expansion_fim_batch}
    \delta(\paramlin, \xinput)(\parampert) := \KL ( p_{\paramlin}(\youtput \vert \xinput) \Vert p_{\paramlin + \parampert}(\youtput \vert \xinput) ) \approx \parampert^\top \fim(\paramlin, \xinput) \parampert + o(\Vert \parampert \Vert^2)
\end{equation}
where $\fim( \paramlin, \xinput) := \jac(\paramlin, \xinput)^\top \noisecov^{-1} \jac(\paramlin, \xinput) = \stdnoise^{-2} \jac(\paramlin, \xinput)^\top \jac(\paramlin, \xinput) \in \R^{\pdim \times \pdim}$ is the empirical Fisher Information Matrix (FIM) of the batch of inputs $\xinput$, computed on the parameters $\paramlin$.

\textbf{Generalization: perturbation of the prediction of a dataset.} Now we can define the influence of a parameter perturbation on the whole training dataset $\dataset$, by generalizing the previous definition:
\begin{displaymath}
    \delta(\paramlin, \dataset)(\parampert) := \dfrac{1}{\ntraintasks} \sum_{i=1}^{\ntraintasks} \delta(\paramlin, \xallcontextinput)(\parampert)
\end{displaymath}
Using \eqref{eq:expansion_fim_batch}, this quantity verifies:
\begin{equation}
    \label{eq:expansion_fim}
     \delta(\paramlin, \dataset)(\parampert) \approx \parampert^\top \left(\dfrac{1}{\ntraintasks} \sum_{i=1}^{\ntraintasks} \fim(\paramlin, \xallcontextinput) \right) \parampert + o(\Vert \parampert \Vert^2)
\end{equation}
which gives a natural definition for the FIM of the whole dataset by analogy with \eqref{eq:expansion_fim_batch}:
\begin{displaymath}
    \fim(\paramlin, \dataset) := \dfrac{1}{\ntraintasks} \sum_{i=1}^{\ntraintasks} \fim(\paramlin, \xallcontextinput) = \dfrac{1}{\ntraintasks \stdnoise^2} \sum_{i=1}^{\ntraintasks} \jac(\paramlin, \xallcontextinput)^\top \jac(\paramlin, \xallcontextinput) \in \R^{\pdim \times \pdim}
\end{displaymath}
The expansion in \eqref{eq:expansion_fim} shows that the FIM of the dataset is a second-order approximation describing the influence of a parameter perturbation over the entire dataset. In particular, the eigenvectors of $\fim(\paramlin, \dataset)$ with the highest eigenvalues are the directions that impact the most the predictions.

\subsection{Computing the top eigenvectors of the Fisher Information Matrix of the dataset in a deep learning context}
\label{app:sketching}

Naively computing the top eigenspace of $\fimd$ requires processing a $\pdim \times \pdim$ matrix, which is intractable in the deep-learning context (where $\pdim$ can surpass $10^6$).
Instead, we decide to use the method used by \citet{scod}, which leverages a low-rank-approximation-based technique (namely matrix sketching) developed by \citet{sketching}.

\paragraph{Sketching the FIM of the dataset} The key idea behind this technique is to build two small random sketches of the FIM, $(\sketchY, \sketchW) := \sketch \left( \fimd \right)$, that with high probability contain enough information to reconstruct the top $\sdim$ eigenvectors of $\fimd$.
The linearity of $\sketch$ simplifies the sketching process by breaking down the computation into individual sketches:
\begin{displaymath}
    (\sketchY, \sketchW) := \sketch (\fimd) = \dfrac{1}{\ntraintasks \stdnoise^2} \sum_{i=1}^\ntraintasks \sketch \left(\jac(\paramlin, \xallcontextinput)^\top \jac(\paramlin, \xallcontextinput)\right) =: \dfrac{1}{\ntraintasks \stdnoise^2} \sum_{i=1}^\ntraintasks (\sketchY^i, \sketchW^i)
\end{displaymath}
In particular, the sketch $(\sketchY, \sketchW)$ can be updated in-place and does not require to store all the individual sketches.
Given a sketch budget $k+l$ (\citet{scod} recommends choosing $k := 2\sdim + 1$ and $l := 4r+3$) and two random normal matrices $\sketchOm \in \R^{k \times \pdim}$ and $\sketchPsi \in \R^{l \times \pdim}$, the random individual sketches $(\sketchW^i, \sketchY^i)$ are defined as:
\begin{displaymath}
\left\{
\begin{array}{rcll}
    \sketchY^i & := & ((\sketchOm  \jac_i^\top) \jac_i)^\top & \in \R^{\pdim \times k} \\
    \sketchW^i & := & (\sketchPsi \jac_i^\top) \jac_i & \in \R^{l \times \pdim}
\end{array}
\right.
\end{displaymath}
where $\jac_i := \jac( \paramlin, \xallcontextinput )$.

\paragraph{Sketch-based computation of the top eigenspace} Once the sketches are computed, the function \Call{FixedRankSymApprox}{} by \citet{sketching} computes the first eigenvectors and eigenvalues of the FIM of the dataset.
Overall, the memory footprint to find the sketches and the top eigenspace is $O(\pdim (\sdim + \ydim \ntraindatapoints))$, where $\sdim$ is the number of queried eigenvectors and $\ntraindatapoints$ is the size of $\xallcontextinput$.
As long as $\sdim \ll \pdim$, this computation is tractable, given that $\ydim \ntraindatapoints \ll \pdim$ is usual deep learning contexts.
Algorithm~\ref{alg:fim_proj} summarizes the process that yields the FIM-based projections via sketching.
We drop the scaling coefficient $\stdnoise^{-2}$ as it doesn't affect the computation, given that we only want orthogonal eigenvectors and eigenvalues.

\begin{algorithm}
\begin{algorithmic}[1]
    \Require $\sdim$ (desired size of the subspace)
    \State $k \gets 2 \sdim + 1$
    \State $l \gets 4 \sdim + 3$
    \State Draw $\sketchOm \in \R^{k \times \pdim}, \sketchPsi \in \R^{l \times \pdim}$, two random normal matrices
    \State Initialize $\sketchY = 0 \in \R^{\pdim \times k}, \sketchW = 0 \in \R^{l \times \pdim}$
    \ForAll{training task $\task$}
        \State $\jac_i \gets \jac(\paramlin, \xallcontextinput)$
        \State $\sketchY \gets \sketchY + 1/\ntraintasks ((\sketchOm \jac_i^\top)\jac_i)^\top$
        \State $\sketchW \gets \sketchW + 1/\ntraintasks (\sketchPsi \jac_i^\top)\jac_i$
    \EndFor
\end{algorithmic}
\caption{Computing the FIM-based projections}
\label{alg:fim_proj}
\end{algorithm}

\subsection{Computational Complexity Analysis}
\label{sec:appendix_complexity}

The computational complexity of our method, \approach{}, scales \textbf{linearly with the number of model parameters ($P$)}, which is comparable to the scaling of MAML. The primary additional costs arise from Gaussian Process (GP) matrix operations, which remain manageable in typical meta-learning settings. Below, we provide a detailed breakdown of the computational costs for both the training and inference phases.

To recap, the variables used in this analysis are defined as:
\begin{itemize}
    \item $P$: The total number of parameters in the neural network.
    \item $K$: The number of context points in a given task.
    \item $N_y$: The output dimensionality of the model.
    \item $r$: The rank of the low-rank covariance approximation, where $r \ll P$.
\end{itemize}

\subsection*{Training Cost}

The training cost is dominated by the computation of the Jacobian and the GP prior covariance. The cost varies depending on the chosen covariance parameterization.

\begin{itemize}
    \item \textbf{\approach{} with Low-Rank Covariance (\approach{}-F and -R):} The primary operations are the GP prior mean computation ($O(N_y K P)$) and the prior covariance computation. The total cost is dominated by the Jacobian-vector products and matrix multiplications involving the rank $r$ factor.
    \[
    \text{Training Cost (Low-Rank)} = O(N_y K r P)
    \]

    \item \textbf{\approach{} with Full Covariance:} Without low-rank approximations, the training cost scales quadratically with the number of parameters $P$ due to the manipulation of the full $P \times P$ covariance matrix.
    \[
    \text{Training Cost (Full)} = O(N_y K P^2)
    \]
\end{itemize}
This comparison highlights the significant efficiency gains from our proposed low-rank parameterizations, making the framework scalable to larger networks.

\subsection*{Inference Cost}

During inference, the cost is determined by the posterior predictive computation, which involves kernel matrix calculations and a matrix inversion. The inference cost is the same for all variants of \approach{}.

The key computational steps are:
\begin{enumerate}
    \item \textbf{Kernel Matrix Computation:} Calculating the kernel matrices $k(x, X)$ and $k(X, X)$ requires Jacobian-vector products, resulting in a cost of $O(N_y^2 K^2 P)$.
    \item \textbf{GP Inverse Computation:} The inversion of the $N_yK \times N_yK$ kernel matrix has a cost of $O((N_yK)^3)$.
\end{enumerate}

Combining these steps, the total inference cost is:
\[
\text{Inference Cost} = O(N_y^2 K^2 P + (N_yK)^3)
\]
In typical meta-learning scenarios, the number of context points $K$ and the output dimension $N_y$ are small, making the $O((N_yK)^3)$ term manageable and often negligible compared to the term that scales with the model size $P$. Furthermore, the kernel computations involving the context set $X$ can be pre-computed and cached, accelerating predictions for multiple query points.

\section{\approach{} as a generalization of ALPaCA}
\label{app:link_with_alpaca}

\paragraph{Restraining the linearization to the last layer} Remember the linear regression of \eqref{eq:linearized_network}, that we obtained by linearizing the network with all its layers.
Let's separate the parameters of the network $\param$ into two: the parameters of all the layers but the last one $\firstparam$, and the parameters of the last layer $\lastparam$: $\param = \firstparam \cup \lastparam$.

We assume that the last layer is dense with biases: we will note $\lastparam^w$ the weight matrix and $\lastparam^b$ the biases of this last layer. $\lastdim$ will stand as the dimension of the activations right before the last layer: in particular, $\lastparam^w \in \R^{\ydim \times \lastdim}$ and $\lastparam^b \in \R^{\ydim}$. $\pdim'$ will stand as the size of $\lastparam$: in our case, $\pdim'=\lastdim \times \ydim + \ydim$. We implicitly vectorize $\lastparam$, such that:
\begin{displaymath}
    \lastparam =
    \begin{pmatrix}
        \vectorized{\lastparam^w} \\
        \lastparam^b
    \end{pmatrix} \in \R^{\lastdim \ydim + \ydim} = \R^{\pdim'}
\end{displaymath}

We now restrain the linear regression to the last layer as follows:
\begin{equation}
\label{eq:reglin_restrained}
    \ysingleoutput = \jac'(\lastparam_0, \psi_{\firstparam}(\xinput)) ( \lastparam - \lastparam_0 )+ \noise
\end{equation}
where:
\begin{itemize}
    \item $\psi_{\firstparam}(\cdot): \R^{\xdim \times \batchsize} \rightarrow \R^{\lastdim \times \batchsize}$ stands for the function that maps the inputs and the activations right before the last layer;
    \item $\jac'(\cdot, \cdot):\R^{\pdim'} \times \R^{\lastdim \times \batchsize} \rightarrow \R^{\ydim \batchsize \times \pdim'}$ stands for the jacobian of the last layer \textit{with respect to the parameters $\lastparam$}. We can write the jacobian in closed-form due to the linearity of the last layer:
    \begin{displaymath}
        \jac'(\lastparam_0, \psi_{\firstparam}(\xinput)) = 
        \begin{pmatrix}
            \psi_{\firstparam}(\xinput)^\top \otimes \mI_{\ydim} & \mI_{\ydim}
        \end{pmatrix}
        \in \R^{\ydim \batchsize \times \pdim'}
    \end{displaymath}
    Note that $\jac'(\lastparam_0, \psi_{\firstparam}(x))$ does not depend on the linearization point $\lastparam_0$ (could be expected, given the linearity of the last layer).
    \item $\lastparam - \lastparam_0 \in \R^{\pdim'}$ is the \textit{correction} to the last parameters. We will note the correction to the weights as $\hat{\lastparam}^w := \lastparam^w - \lastparam_0^w$ and the correction to the bias as $\hat{\lastparam}^b := \lastparam^b - \lastparam_0^b$.
\end{itemize}

Using Kronecker's product identities, the linear regression in \eqref{eq:reglin_restrained} can be rewritten:
\begin{equation}
    \label{eq:reglin_last}
    \ysingleoutput = \hat{\lastparam}^w \psi_{\firstparam}(\xinput) + \hat{\lastparam}^b + \noise
\end{equation}

Also, as a side note, another way of getting this linear regression (\eqref{eq:reglin_last}) is to rewrite the initial linearization of Section Background, but with respect to $\lastparam$ only:
\begin{displaymath}
    \begin{array}{rcl}
        \modelsingleinput (\firstparam \cup \lastparam, \xsingleinput_t) - \modelsingleinput (\firstparam \cup \lastparam_0, \xsingleinput_t) & = & \lastparam^w \psi_{\firstparam}(x) + \lastparam^b - (\lastparam_0^w \psi_{\firstparam}(x) + \lastparam_0^b) \\
         & = & (\lastparam^w - \lastparam_0^w) \psi_{\firstparam}(x) + (\lastparam^b - \lastparam_0^b) \\
         & = & \hat{\lastparam}^w \psi_{\firstparam}(x) + \hat{\lastparam}^b
    \end{array}
\end{displaymath}
Doing it this way has the benefit to show that the linearization is exact when restricted to the last layer.
For non-linear neural networks, the linearization is always an approximation.

\paragraph{Adapting \approach{} to the last layer}
Just like what we did in the general case, we use the GP theory to make Bayesian inference on this linear regression.
The new parameters of the GP $\distparam$ are now $\xi = (\firstparam, \priormean, \priorcov)$, where $\priormean$ and $\priorcov$ are the parameters of the Gaussian prior over $\hat{\lastparam}$.
Note how $\lastparam_0$ have disappeared from $\xi$: contrary to the general case, the linearization point is not optimized, as it does not impact the computation.
Also note that $\firstparam$ has replaced $\lastparam_0$, as it parameterizes the feature map.

\paragraph{Comparison with ALPaCA}
In the ALPaCA setting, the linear regression is:
\begin{equation}
    \label{eq:reglin_alpaca}
    \ysingleoutput_a = \hat{\lastparam}^w \psi_{\firstparam}(\xinput) + \noise
\end{equation}
Note that $\hat{\lastparam}^w$ still plays the same role in the linear regression, but is not any \textit{correction} anymore.
Subtracting \ref{eq:reglin_last} from \ref{eq:reglin_alpaca} yields:
\begin{displaymath}
    \ysingleoutput - \ysingleoutput_a = \hat{\lastparam}^b
\end{displaymath}
\approach{} restricted to the last layer is closely related to ALPaCA, in that they both perform a linear regression with the same kernel: the only difference lies in the additional bias term that is not considered in ALPaCA.
Thus, we can think of \approach{} as a generalization of ALPaCA to all the layers of the network.

\revise{
\section{Comparison with Deep Kernel Learning}
\label{app:dkt_comparison}

Table~\ref{tab:dkt_vs_ours} summarizes the key representational differences between Deep Kernel Transfer (DKT) and our method.

\begin{table}[h]
\centering
\small
\begin{tabular}{l|p{6cm}|p{7.5cm}}
\toprule
\textbf{Aspect} & \textbf{Meta Deep Kernel Learning} & \textbf{Ours} \\
\midrule
\textbf{Kernel} & Over feature \emph{outputs}: $k(h_\theta(x_1), h_\theta(x_2))$ & In \emph{weight space} via NTK: $k_\Sigma(x_1, x_2) = J(x_1)^\top \Sigma J(x_2)$ \\
\midrule
\textbf{Meta-learned} & Feature extractor weights $\theta$ & Prior distribution $(\mu, \Sigma)$ over network weights \\
\midrule
\textbf{Uncertainty} & Input-level (distance in feature space) & Task-level (prior predictive over context) \\
\bottomrule
\end{tabular}
\caption{Comparison of representational differences between DKT and \approach{}.}
\label{tab:dkt_vs_ours}
\end{table}
}

\section{Additional results}

\subsection{Additional results (single-cluster case)}
\label{app:additional-single}

\begin{figure}
     \centering
     \begin{subfigure}[t]{0.3\textwidth}
         \centering
         \includegraphics[width=\textwidth]{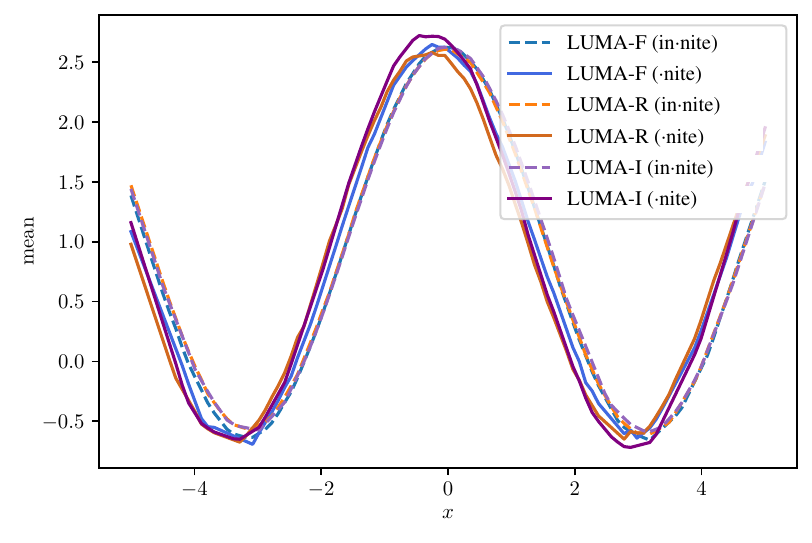}
         \caption{Mean $\mathbb{E}(f)$ (\trainingid{}, \trainingrandom, \trainingfim)}
         \label{fig:single-mean}
     \end{subfigure}
     \hfill
     \begin{subfigure}[t]{0.3\textwidth}
         \centering
         \includegraphics[width=\textwidth]{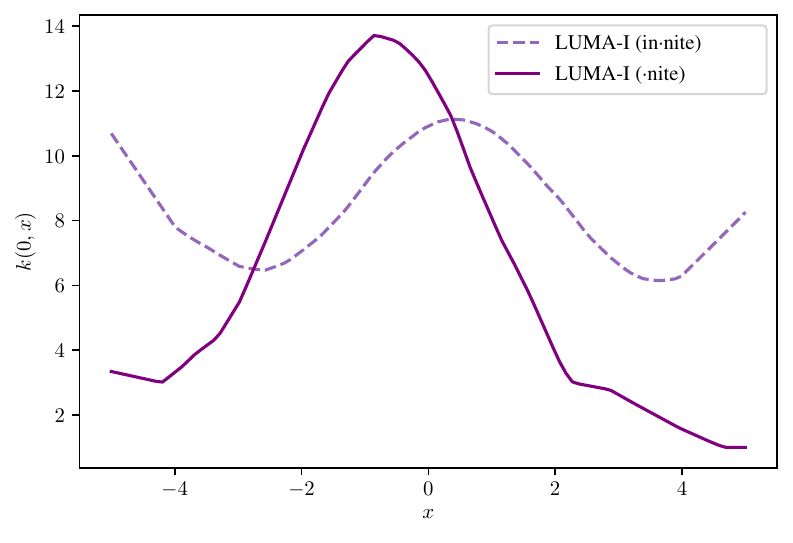}
         \caption{Covariance $\text{cov}(f(0), f(x))$ (\trainingid)}
         \label{fig:single-cov-id}
     \end{subfigure}
     \hfill
     \begin{subfigure}[t]{0.3\textwidth}
         \centering         \includegraphics[width=\textwidth]{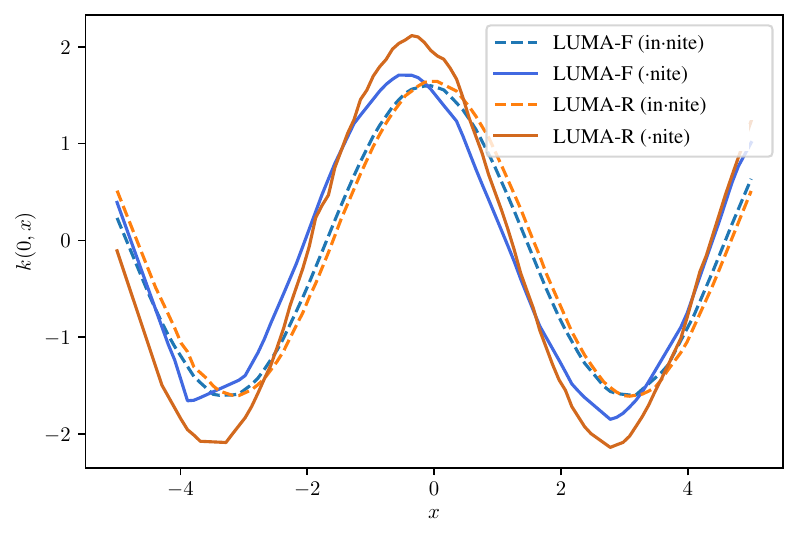}
         \caption{Covariance $\text{cov}(f(0), f(x))$ (\trainingrandom{}, \trainingfim)}
         \label{fig:single-cov}
     \end{subfigure}
    \caption{Mean (\ref{fig:single-mean}) and covariance  (\ref{fig:single-cov-id}, 
\ref{fig:single-cov}) functions of $\distparam$ after different meta-trainings on the sine cluster (\trainingid, \trainingrandom{} and \trainingfim, with a finite or infinite dataset). Note the scale, and how \trainingid{} has a less accurate covariance function that \trainingrandom{} and \trainingfim.}
    \label{fig:single-gp}
\end{figure}

\paragraph{Quality of the priors}
To qualitatively analyze the prior after meta-training on the sines, we plot the mean functions and the covariance functions of the resulting GP, that is $\distparam$ (Figure~\ref{fig:single-gp}).
All the trainings yield a similar mean for $\distparam$ (that is a cosine with amplitude 1.5 and offset 1) (Figure~\ref{fig:single-mean}), which is close to the theoretical mean of $\dist$ (a cosine with amplitude 2.5 and offset 1).
The covariance function of \trainingrandom{} and \trainingfim{} (Figure~\ref{fig:single-cov}) resembles what we would expect for $\dist$ (e.g., periodicity, negative correlation between 0 and $\pi$, etc.), but it is not the case for \trainingid{} (Figure~\ref{fig:single-cov-id}).
This empirical analysis confirms the quantitative comparison in terms of OoD detection and prediction performance carried in Section Numerical Analysis.

\begin{figure}[t]
    \centering
    \begin{subfigure}[t]{0.27\textwidth}
        \centering
        \includegraphics[width=\textwidth]{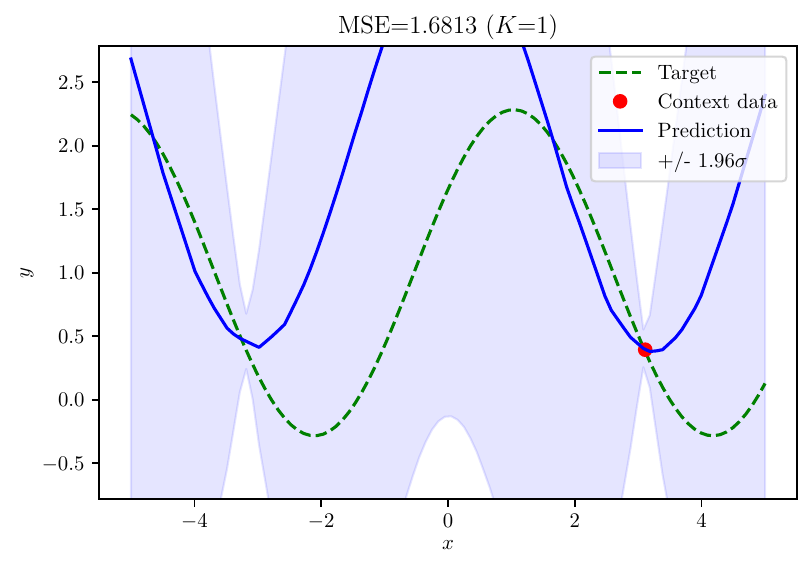}
        \includegraphics[width=\textwidth]{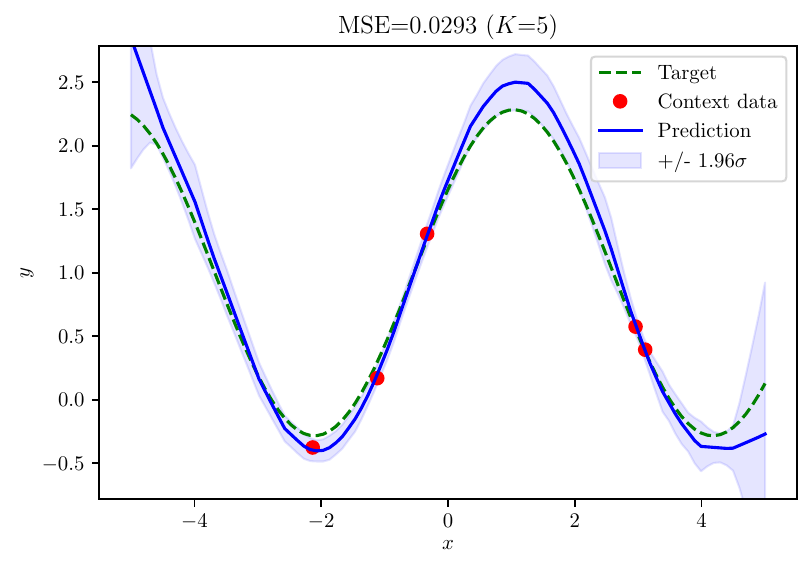}
        \includegraphics[width=\textwidth]{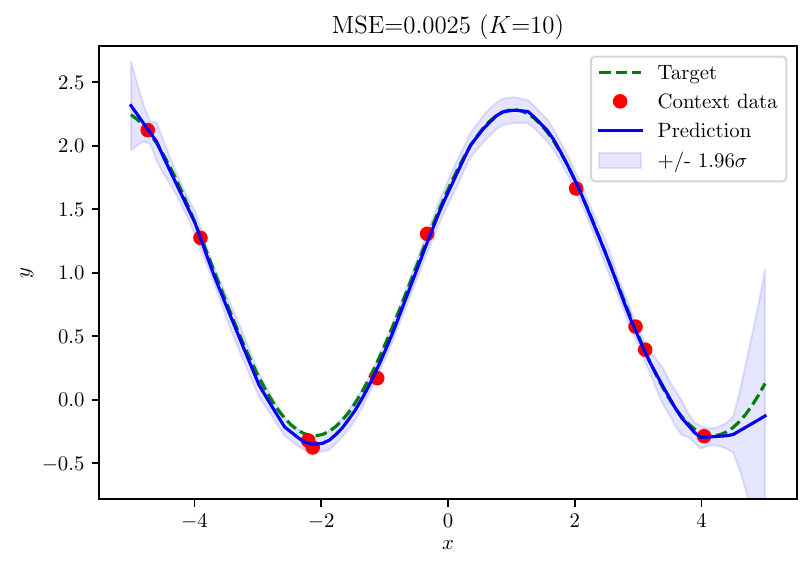}
        \caption{\trainingid{} (inf)}
    \end{subfigure}
    \hfill
    \begin{subfigure}[t]{0.27\textwidth}
        \centering
        \includegraphics[width=\textwidth]{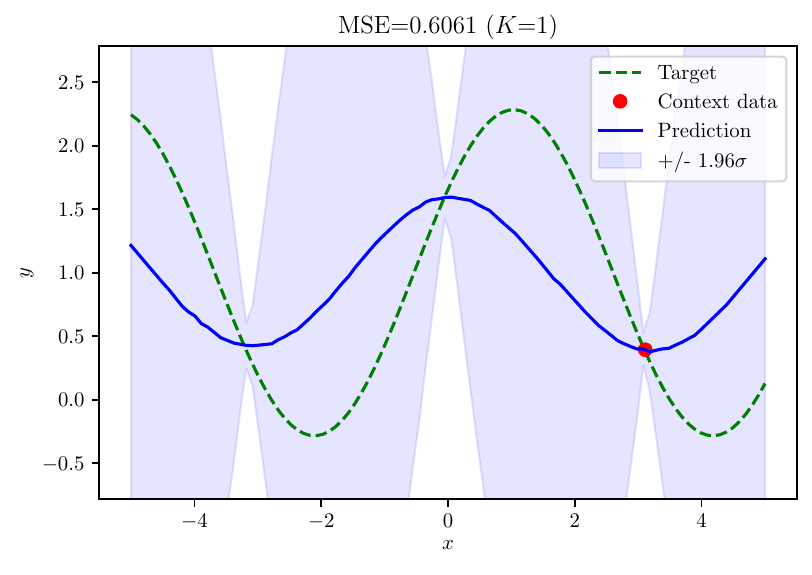}
        \includegraphics[width=\textwidth]{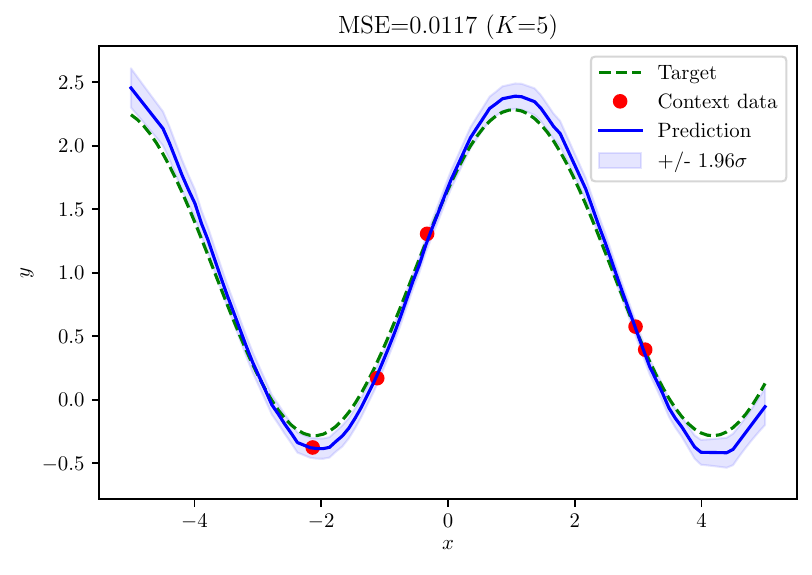}
        \includegraphics[width=\textwidth]{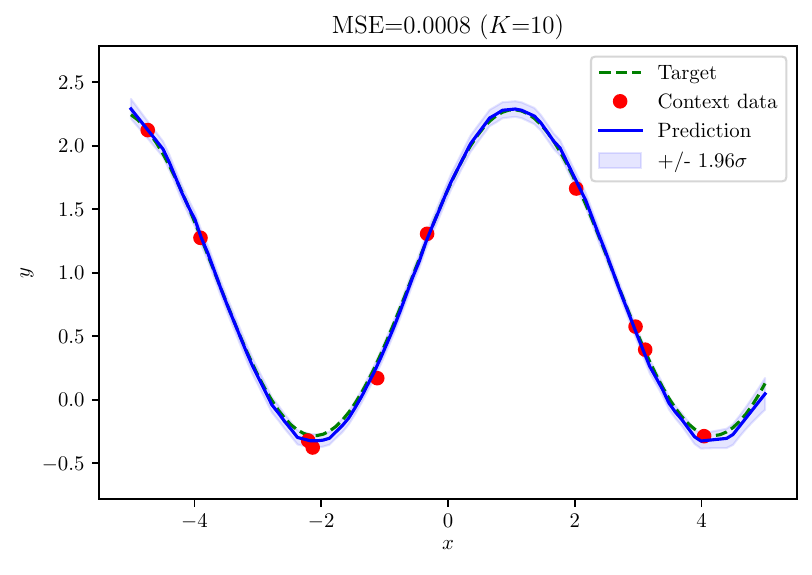}
        \caption{\trainingrandom{} (inf)}
    \end{subfigure}
    \hfill
    \begin{subfigure}[t]{0.27\textwidth}
        \centering
        \includegraphics[width=\textwidth]{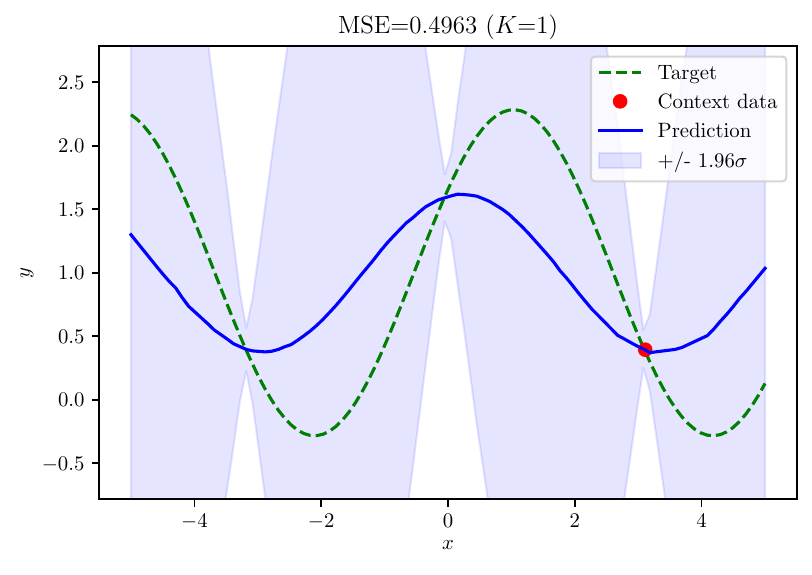}
        \includegraphics[width=\textwidth]{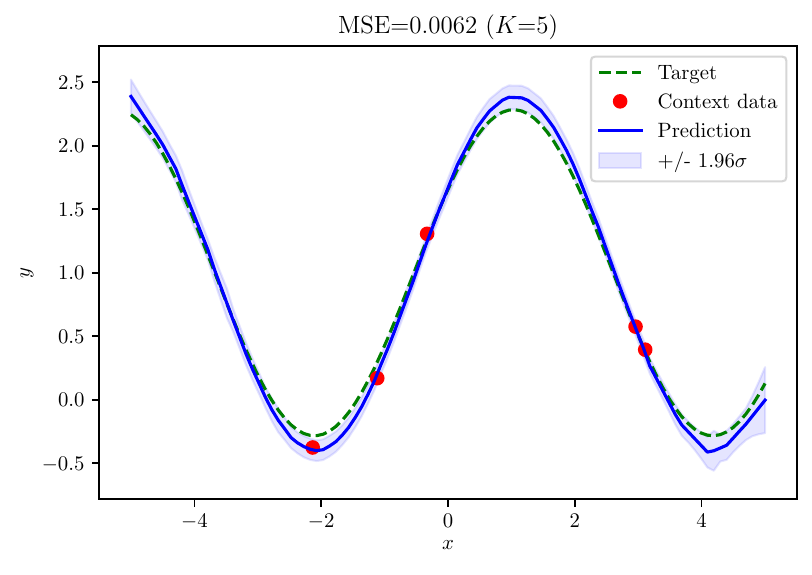}
        \includegraphics[width=\textwidth]{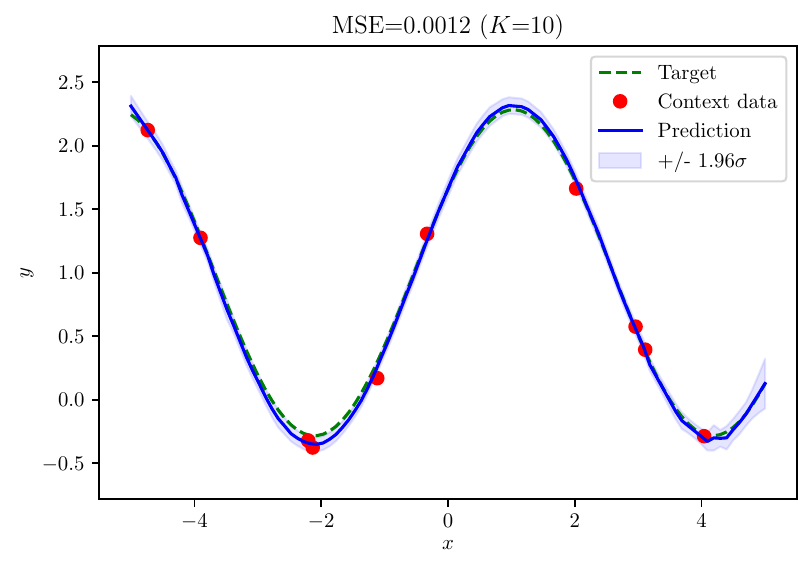}
        \caption{\trainingfim{} (inf)}
    \end{subfigure}
    \hfill
    \begin{subfigure}[t]{0.27\textwidth}
        \centering
        \includegraphics[width=\textwidth]{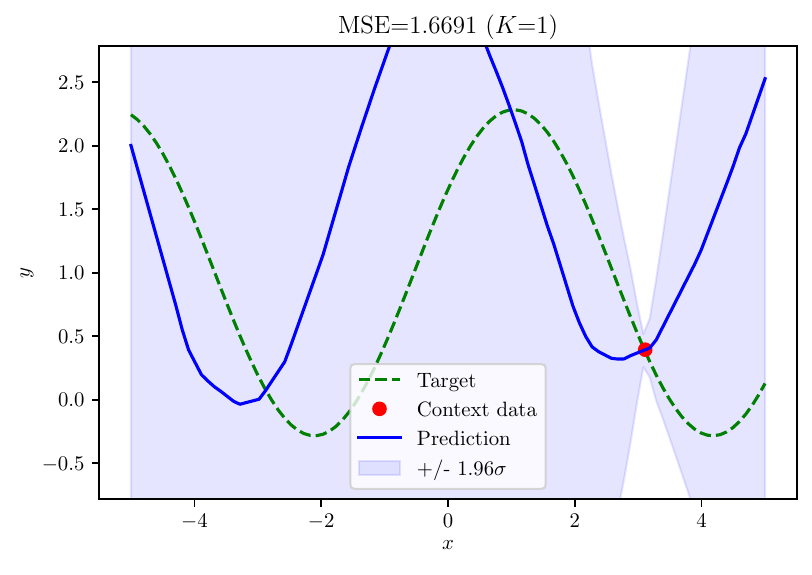}
        \includegraphics[width=\textwidth]{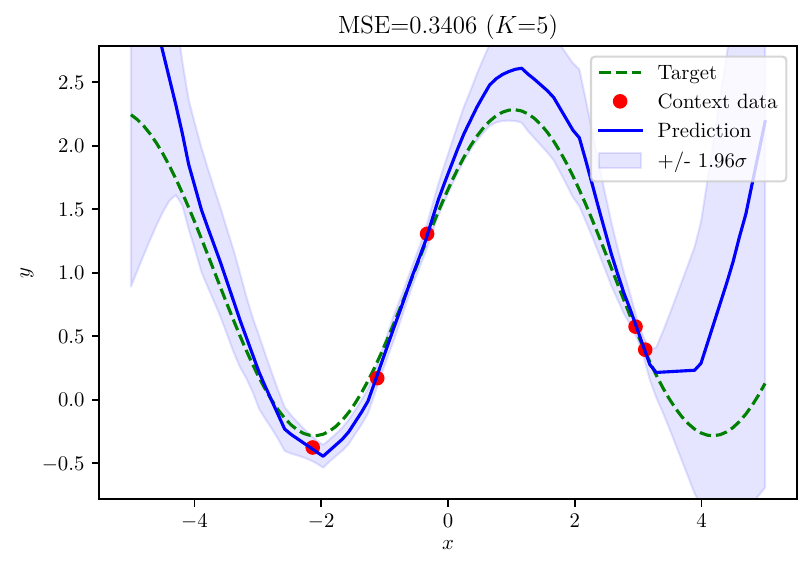}
        \includegraphics[width=\textwidth]{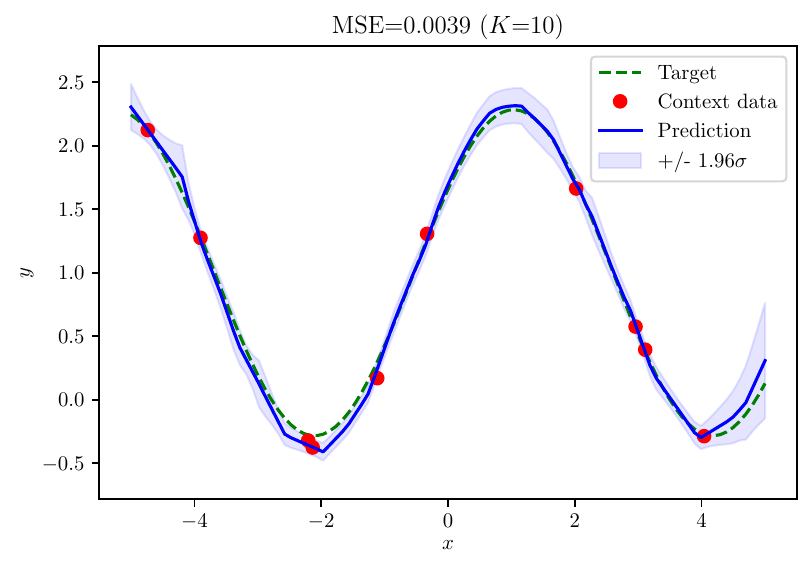}
        \caption{\trainingid{} (fin)}
    \end{subfigure}
    \hfill
    \begin{subfigure}[t]{0.27\textwidth}
        \centering
        \includegraphics[width=\textwidth]{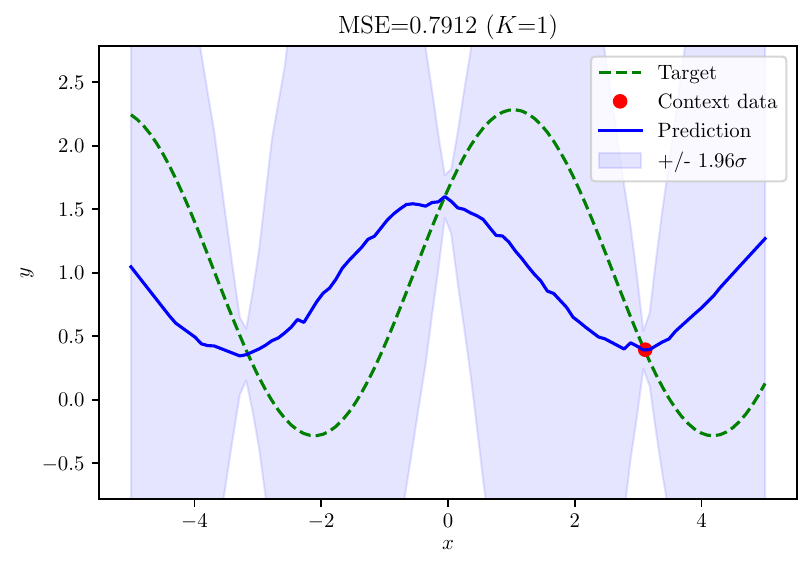}
        \includegraphics[width=\textwidth]{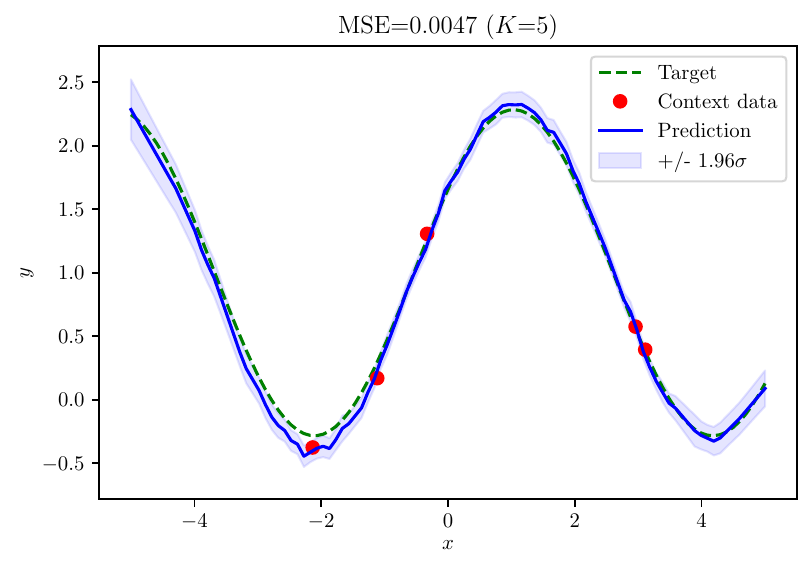}
        \includegraphics[width=\textwidth]{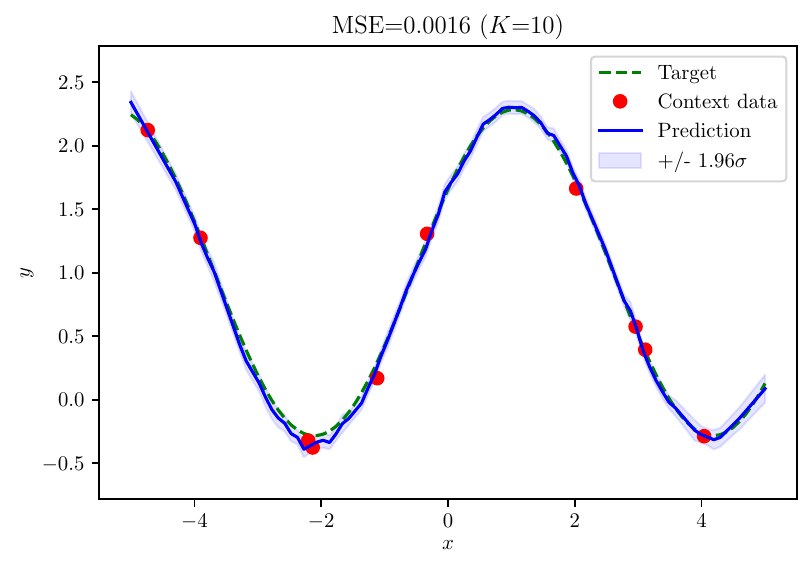}
        \caption{\trainingrandom{} (fin)}
    \end{subfigure}
    \hfill
    \begin{subfigure}[t]{0.27\textwidth}
        \centering
        \includegraphics[width=\textwidth]{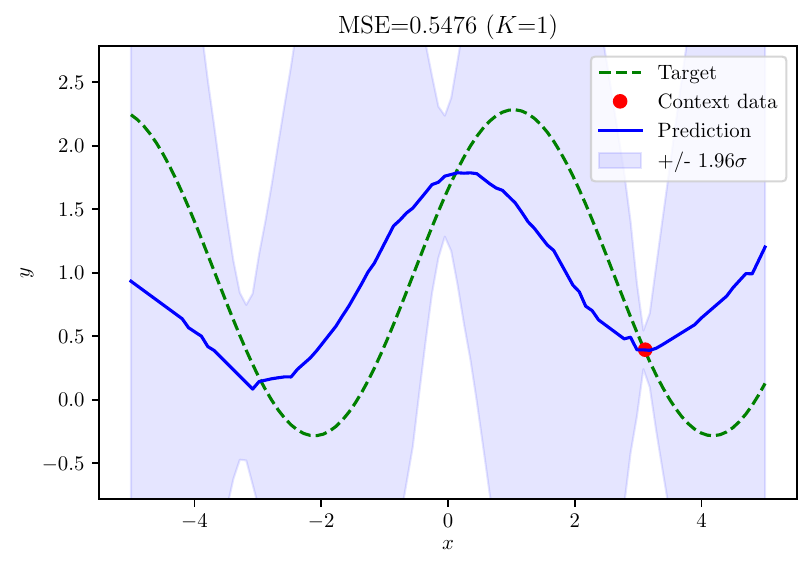}
        \includegraphics[width=\textwidth]{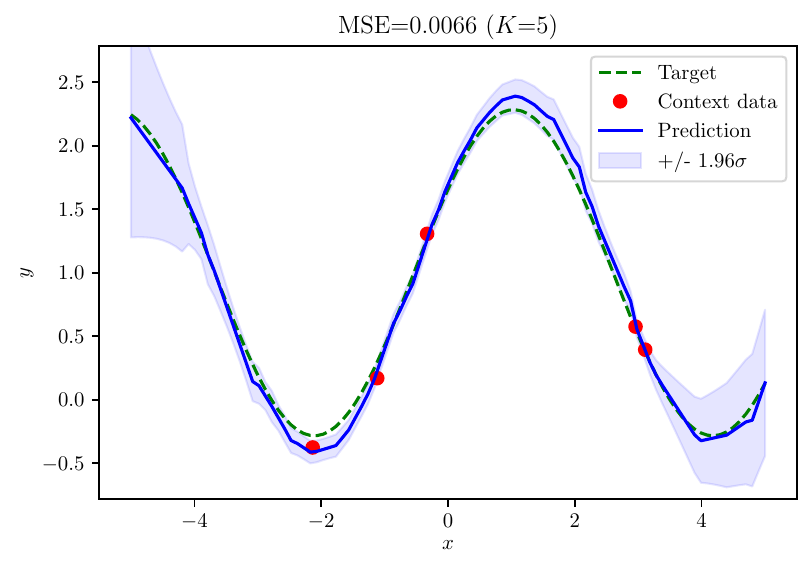}
        \includegraphics[width=\textwidth]{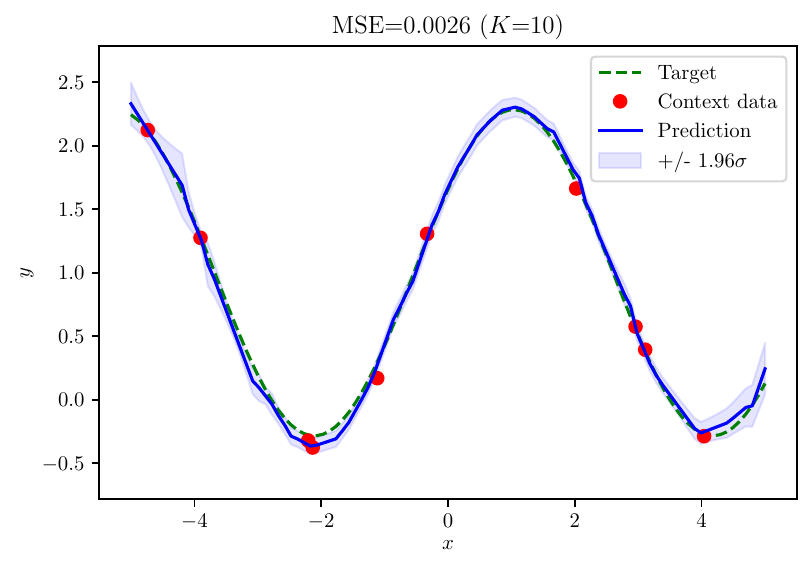}
        \caption{\trainingfim{} (fin)}
    \end{subfigure}
    \caption{Example of predictions for a varying number of context inputs $\batchsize$, after different meta-trainings on the sine cluster (\trainingid, \trainingrandom{} and \trainingfim, in the infinite and finite case). Standard deviation is from the posterior predictive distribution. Note how \trainingrandom{} and \trainingfim{} perform better than \trainingid{} when it comes to reconstructing the sine with a smaller amount of context inputs.}
    \label{fig:single-predictions-full}
\end{figure}

\paragraph{Examples of predictions} Figure~\ref{fig:single-predictions-full} summarizes the predictions of the model meta-trained on sines, for a varying number of context inputs $\batchsize$, breaking down all the different cases of training.

\subsection{Additional results (multi-cluster case)}
\label{app:additional-multi}

\begin{figure}
     \centering
     \begin{subfigure}[t]{0.3\textwidth}
         \centering
         \includegraphics[width=\textwidth]{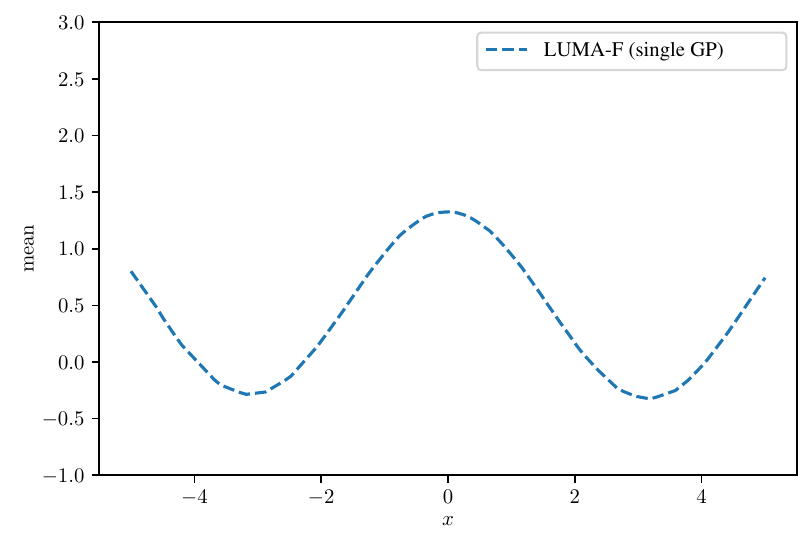}
         \caption{Mean function $\mathbb{E}(f)$ of the GP (\trainingfim{} with a single GP).}
     \end{subfigure}
     \hfill
     \begin{subfigure}[t]{0.3\textwidth}
         \centering
         \includegraphics[width=\textwidth]{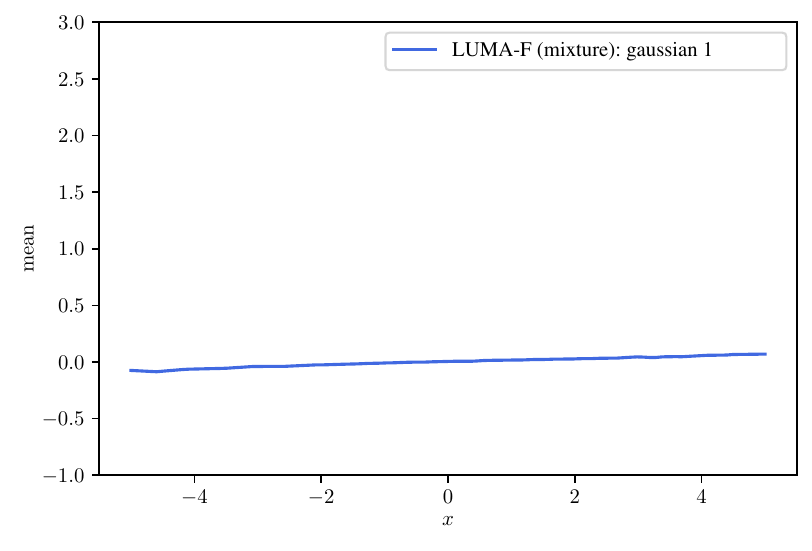}
         \caption{Mean function $\mathbb{E}(f)$ of the first GP of the mixture (\trainingfim{} with a mixture of GPs).}
         \label{fig:multi-mean-gauss1}
     \end{subfigure}
     \hfill
     \begin{subfigure}[t]{0.3\textwidth}
         \centering
         \includegraphics[width=\textwidth]{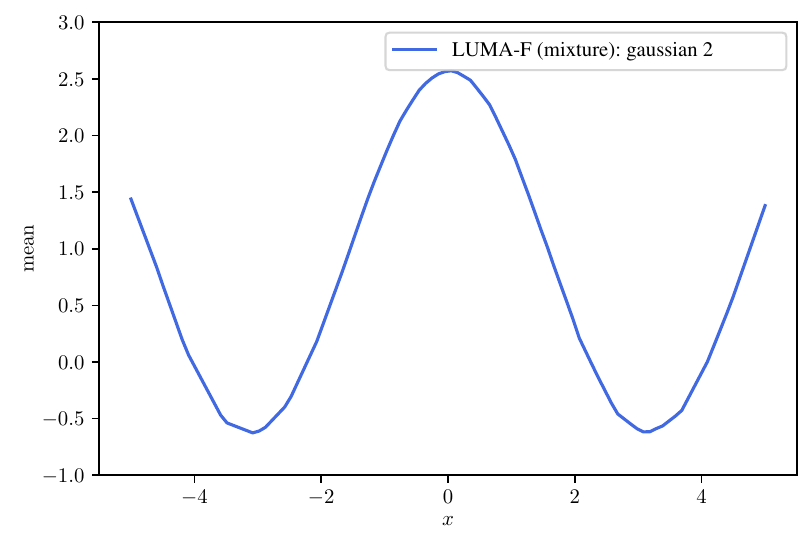}
         \caption{Mean function $\mathbb{E}(f)$ of the second GP of the mixture (\trainingfim{} with a mixture of GPs).}
     \end{subfigure}
    \caption{Mean functions of the GP (when learning with a GP) / the GPs composing the mixture (when learning a mixture of GPs), after training on both the sine and line cluster with \trainingfim. Note how the mean of the single GP is intermediate between the ones of the mixture.}
    \label{fig:multi-mean}
\end{figure}

\begin{figure}[t]
     \centering
     \begin{subfigure}[t]{0.3\textwidth}
         \centering
         \includegraphics[width=\textwidth]{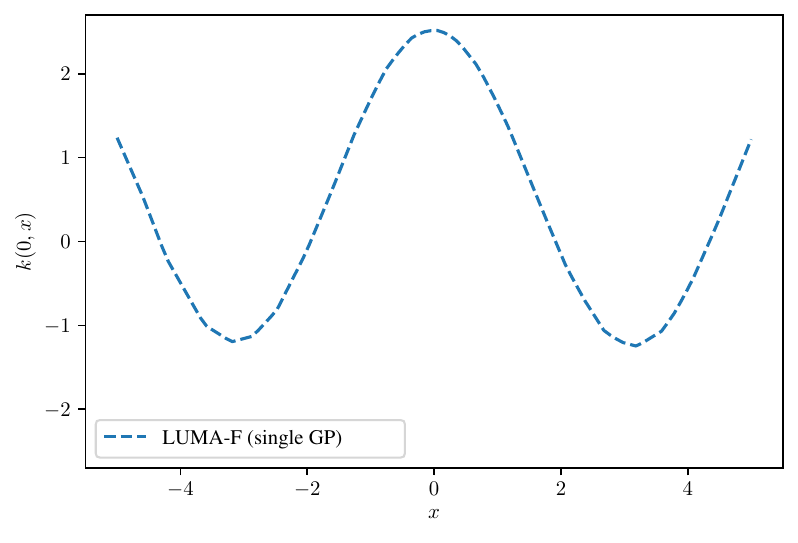}
         \caption{Covariance function $\text{cov}(f(0), f(x))$ of the GP (\trainingfim{} with a single GP).}
     \end{subfigure}
     \hfill
     \begin{subfigure}[t]{0.3\textwidth}
         \centering
         \includegraphics[width=\textwidth]{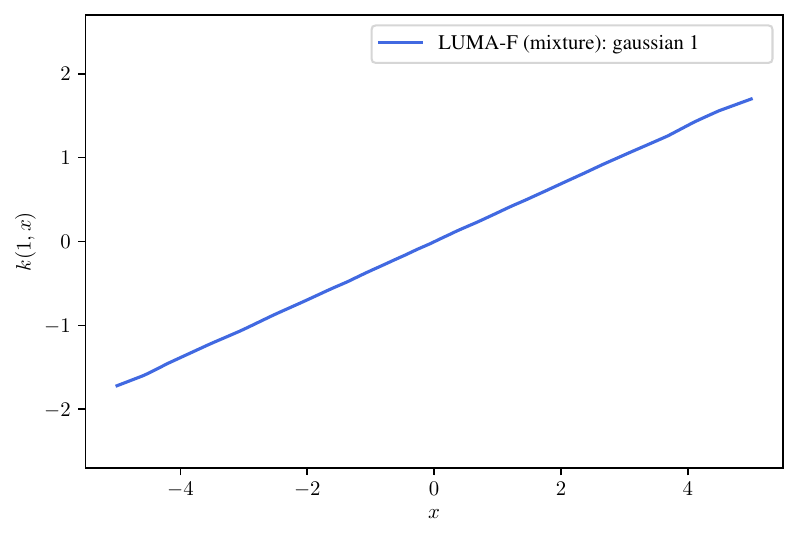}
         \caption{Covariance function $\text{cov}(f(1), f(x))$ of the first GP of the mixture (\trainingfim{} with a mixture of GPs)}
         \label{fig:multi-cov-gauss1}
     \end{subfigure}
     \hfill
     \begin{subfigure}[t]{0.3\textwidth}
         \centering
         \includegraphics[width=\textwidth]{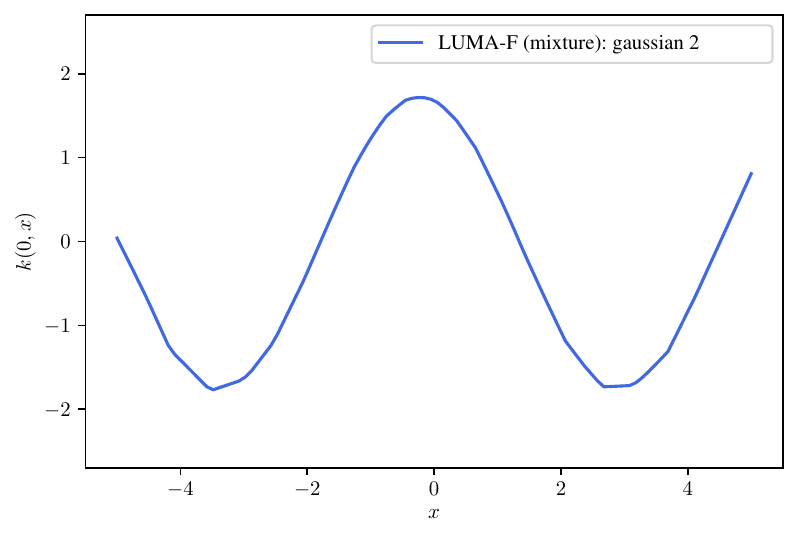}
         \caption{Covariance function $\text{cov}(f(0), f(x))$ of the second GP of the mixture (\trainingfim{} with a mixture of GPs)}
     \end{subfigure}
    \caption{Covariance functions of the GP (when learning with a GP) / the GPs composing the mixture (when learning a mixture of GPs), after training on both the sine and line cluster with \trainingfim. Note how the covariance of the single GP is not accurate for any of the two clusters.}
    \label{fig:multi-cov}
\end{figure}

\paragraph{Quality of the priors} Figure~\ref{fig:multi-mean} and Figure~\ref{fig:multi-cov} show the mean and covariance functions of the GP (when training with a single GP), and the mean and covariance functions of the GPs composing the mixture(when training with a mixture of GPs).
We note that in the case of the mixture, both of the Gaussians composing the mixture have correctly captured the common features shared by each of the clusters (respectively the linear and the sine cluster). For instance, the mean of the line cluster is the zero function, which matches Figure~\ref{fig:multi-mean-gauss1}, and the correlation between $x=1$ and the other inputs is correctly rendered for linear tasks (Figure~\ref{fig:multi-cov-gauss1}).

When learning with a single GP, the learnt mean and covariance do not match any of the two clusters. For example, the mean has an intermediate offset between the offset of the sine cluster and the line cluster.
This empirical analysis comforts the conclusions of Section Numerical Analysis: the mixture model yields better results than the single GP case.

\subsection{\approach{} yields effective predictions on large-scale vision problems}
\label{app:deep}
We consider a vision meta-learning problem from \citet{what-matters}, Shapenet1D, aiming to predict object orientations in space.
In this problem, each task consists of a different object of which we want to predict the orientation.
For each task, the context data consists of some images of the same object, but with different orientations; the query inputs are other images of the same object, with unknown orientations.
Details on the problems and the datasets can be found in Appendix \ref{app:problem-details}.

We train a deep learning model on Shapenet1D, and we compare the performances on the test set between \trainingid, \trainingrandom{} and \trainingfim{} (Figure~\ref{fig:vision-performance}). More training and test details can be found in the Appendix~\ref{app:train-details-vision}.

Both \trainingid{} and \trainingfim{} yield better performances than MAML, achieving low angle errors.
\trainingrandom{} however gives poor results, worse than that of MAML.

In terms of the trade-off of Section Methods, our conclusion from small networks does not scale up to deep models.
Here, the loss of valuable features is perceptible (that is what happens with \trainingrandom, when the randomness of the directions may drop such features), and not learning a rich prior over the weights is not burdensome (\trainingid).
However, \trainingfim{} plays a role of compromise, by learning the prior covariance while keeping the few important features of the jacobian, as it gives comparable results to \trainingid.

\section{Details on the regression problems}
\label{app:problem-details}

\subsection{Simple regression problems}
Our simple regression problems are inspired by \citet{mmaml}'s work.
They consists of three clusters of different types of tasks, and varying offset.
The first cluster consists of sines with constant frequency and offset, but with a varying amplitude and phase:
\begin{displaymath}
    \left\{ x \mapsto A \sin (x + \varphi) + 1 \vert A \in [0.1,5], \varphi \in [0, \pi] \right\}
\end{displaymath}
The second cluster consists of lines with a varying slope and no offset:
\begin{displaymath}
    \left\{ x \mapsto a x \vert a \in [-1, 1] \right\}
\end{displaymath}
The last cluster consists of quadratic functions, with a varying quadratic coefficient and phase, and with a constant offset:
\begin{displaymath}
    \left\{ x \mapsto a (x - \varphi)^2 + 0.5 \vert a \in [-0.2, 0.2], \varphi \in [-2, 2] \right\}
\end{displaymath}
In all these clusters, we add an artificial Gaussian noise on the context observations $\normal(0,~0.05)$.
The query datapoints remain noiseless, to remain coherent with the assumption from Section Background borrowed from \citet{rasmussen}.

\subsection{Vision problem}
\begin{figure}
     \centering
     \begin{subfigure}[t]{0.24\textwidth}
         \centering
         \includegraphics[width=\textwidth]{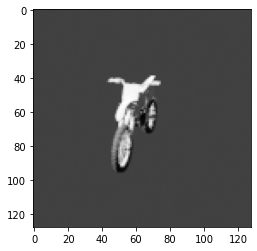}
         \caption{Angle: 4°}
         \includegraphics[width=\textwidth]{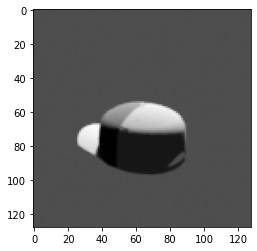}
         \caption{Angle: 26°}
     \end{subfigure}
     \hfill
     \begin{subfigure}[t]{0.24\textwidth}
         \centering
         \includegraphics[width=\textwidth]{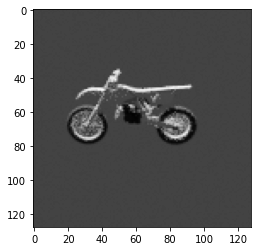}
         \caption{Angle: 78°}
         \includegraphics[width=\textwidth]{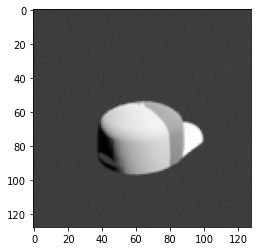}
         \caption{Angle: 130°}
     \end{subfigure}
     \hfill
     \begin{subfigure}[t]{0.24\textwidth}
         \centering
         \includegraphics[width=\textwidth]{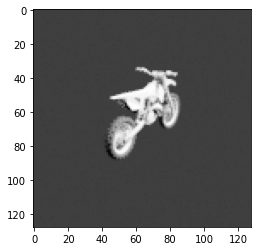}
         \caption{Angle: 202°}
         \includegraphics[width=\textwidth]{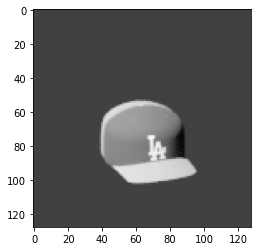}
         \caption{Angle: 243°}
     \end{subfigure}
     \hfill
     \begin{subfigure}[t]{0.24\textwidth}
         \centering
         \includegraphics[width=\textwidth]{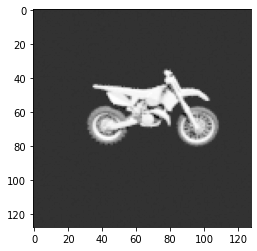}
         \caption{Angle: 261°}
         \includegraphics[width=\textwidth]{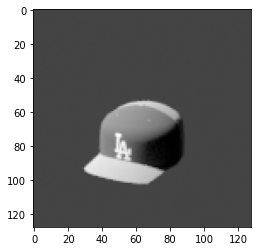}
         \caption{Angle: 285°}
     \end{subfigure}
    \caption{Examples of images and ground-truth angles from the Shapenet1D dataset \citep{what-matters}}
    \label{fig:example_orientations}
\end{figure}

We consider the meta-learning vision, regression problem recently created by \citet{what-matters} (namely Shapenet1D), that consists in estimating objects orientation from images.
The objects have a vertical angular degree of freedom, and Figure~\ref{fig:example_orientations} shows an example of such objects in different positions.

We use the same training and test datasets as \citet{what-matters}, with no kind of augmentation (in particular, no artificial noise on the ground-truth angles).
In particular, we use the same intra-category (IC) evaluation dataset (that is, objects from the same categories as the objects used for training) and cross-category (CC) evaluation dataset (that is, objects from different categories as the ones used for training).

\revise{
\section{Full Results with Confidence Intervals}
\label{app:full_results}

We report the full results with 95\% confidence intervals for all experiments.

\begin{table}[h]
\centering
\setlength{\tabcolsep}{0.8mm}
\footnotesize
\begin{tabular}{l|cc}
\toprule
~ & \textbf{Unimodal (Infinite)} & \textbf{Unimodal (Finite)} \\
\textbf{$\batchsize$} & \textbf{5 / 10} & \textbf{5 / 10} \\
\midrule
CNP & 0.0485$\pm$0.0077 / 0.0189$\pm$0.0030 & 1.9816$\pm$0.1902 / 1.5777$\pm$0.1806 \\
TNP-D & 0.1324$\pm$0.1045 / 0.0186$\pm$0.0014 & 2.5239$\pm$0.2130 / 1.6384$\pm$0.1461 \\
DKT & 3.2730$\pm$0.4264 / 0.2122$\pm$0.0524 & 3.4938$\pm$0.3593 / 0.3008$\pm$0.1193 \\
\textbf{Ours} & \textbf{0.0026$\pm$0.0002} / \textbf{0.0015$\pm$0.0001} & \textbf{0.0204$\pm$0.0018} / \textbf{0.0105$\pm$0.0010} \\
\bottomrule
\end{tabular}
\caption{Unimodal regression MSE with 95\% CIs. Mean $\pm$ CI reported for $\batchsize$=5/10.}
\label{tab:full_results_ci}
\end{table}

\begin{table}[h]
\centering
\setlength{\tabcolsep}{0.8mm}
\footnotesize
\begin{tabular}{l|cc}
\toprule
~ & \multicolumn{2}{c}{\textbf{Multimodal}} \\
\textbf{$\batchsize$} & \textbf{5} & \textbf{10} \\
\midrule
CNP & 0.1402 $\pm$ 0.0490 & 0.0311 $\pm$ 0.0083 \\
TNP-D & 0.1253 $\pm$ 0.0517 & 0.0196 $\pm$ 0.0020 \\
DKT & 2.4779 $\pm$ 0.3681 & 0.1760 $\pm$ 0.0525 \\
Ours (Single) & 0.0454 $\pm$ 0.0084 & 0.0027 $\pm$ 0.0003 \\
\textbf{Ours (Mixt)} & \textbf{0.0024 $\pm$ 0.0002} & \textbf{0.0012 $\pm$ 0.0001} \\
\bottomrule
\end{tabular}
\caption{Multimodal regression MSE with 95\% CIs. \approach{} (Mixture) achieves the lowest error.}
\label{tab:mse_comparison_multimodal}
\end{table}

\begin{table}[h]
\centering
\setlength{\tabcolsep}{0.8mm}
\footnotesize
\begin{tabular}{l|ccc}
\toprule
~ & \multicolumn{3}{c}{\textbf{Vision Regression Tasks (Angular Error $^\circ$)}} \\
\textbf{$\batchsize$} & \textbf{5} & \textbf{10} & \textbf{15} \\
\midrule
CNP & 22.31 $\pm$ 1.930 & 19.70 $\pm$ 1.250 & 21.00 $\pm$ 2.310 \\
TNP-D & 88.74 $\pm$ 1.830 & 89.75 $\pm$ 1.760 & 89.96 $\pm$ 1.660 \\
DKT & 21.32 $\pm$ 1.840 & \textbf{3.920 $\pm$ 0.4600} & \textbf{1.790 $\pm$ 0.1500} \\
\textbf{Ours} & \textbf{18.94 $\pm$ 3.280} & 7.684 $\pm$ 1.150 & 5.157 $\pm$ 0.9000 \\
\bottomrule
\end{tabular}
\caption{Vision regression (angular error) with 95\% CIs. DKT excels with sufficient context ($\batchsize \geq 10$), while \approach{} is relatively more stable in the extreme low-data regime ($\batchsize=5$).}
\label{tab:vision_full_ci}
\end{table}
}

\section{Training and test details for the single cluster case}
\label{app:train-details-single}

\subsection{Training details of \approach{} (single cluster case)}
The model is a neural network with $2$ hidden layers, $40$ neurons each, with a ReLU non-linearity.
In our single-cluster experiment, the cluster is the sine cluster (see Appendix~\ref{app:problem-details}).

In the case where there are an infinite number of available sine tasks during training (\textit{ie} $\ntraintasks=+\infty$), the training is performed with $\ntasksperepoch=24$ tasks per epoch, and at each epoch the context inputs are randomly drawn from $[-5, 5]$ (which means that $\ntraindatapoints=+\infty$).
In the case where we restrict the available tasks to a finite number, we randomly choose $\ntraintasks=10$ tasks and $\ntraindatapoints=50$ context datapoints per task before training (they are shared among all the ``finite'' trainings) and perform the trainings with $\ntasksperepoch=6$ tasks per epoch.

For all the trainings, the number of context inputs seen during training is $\batchsize=10$.

For all trainings, we train \approach{} on $60,000$ epochs.
When dealing with the \trainingfim{} case, we allow half of the epochs ($30,000$) to the training \textit{before} finding the intermediate $\paramlin$ and $\priormean$ (see Algorithm~\ref{alg:meta_training_learnt_cov}), and the other half \textit{after}.

In the \trainingfim{} case with infinite available tasks, a finite number of tasks is needed to compute the FIM: thus we build an artificial finite dataset of $\ntraintasks=100$ and $\ntraindatapoints=\pdim$ (arbitrary, but chosen so that $\jac(\paramlin, \xallcontextinput) \jac(\paramlin, \xallcontextinput)^\top$ gets a chance to be full-rank \textit{ie} contain as much information as possible), that is used only for that computation.

In the \trainingfim{} and the \trainingrandom{} cases, the subspace size is $\sdim=10$.

For all trainings, the meta-optimizer is Adam with an initial learning rate of $0.001$.
The noise $\stdnoise=0.05$, equal to the noise added to the context data.
We compute the NLL using \citet{rasmussen}'s implementation.

\subsection{Training details of MAML (single cluster case)}
We train MAML baselines to compare our results with that of MAML.
A large part of these hyperparameters is directly inspired from \citet{maml}'s work.

The model is a neural network with $2$ hidden layers, $40$ neurons each, with a ReLU non-linearity.
In our single-cluster experiment, the cluster is the sine cluster (see Appendix~\ref{app:problem-details}).

In the case where there are an infinite number of available sine tasks during training (\textit{ie} $\ntraintasks=+\infty$), the training is performed with $\ntasksperepoch=24$ tasks per epoch, and at each epoch the context and query inputs are randomly drawn from $[-5, 5]$.
In the case where we restrict the available tasks to a finite number, we randomly choose $\ntraintasks=10$ tasks and $\ntraindatapoints=50$ datapoints per task (used for both context and query batches) before training (they are shared among all the ``finite'' trainings) and perform the trainings with $\ntasksperepoch=6$ tasks per epoch.

For all trainings, the number of context datapoints is $\batchsize=10$, and the number of query datapoints is $L=10$.
We meta-train for $70,000$ epochs.

For all trainings, the meta-optimizer is Adam with an initial learning rate of $0.001$.
The inner learning-rate is kept constant, at $0.001$.
The number of inner updates is $5$ during training, and $10$ at test time.

\subsection{Test details (single-cluster case)}

For the OoD detection evaluation, we plot the AUC as a function of the number of the context inputs $\batchsize$.
The AUC is computed using the NLL of the context inputs with respect to $\distparam$ (our uncertainty metric).
The true-positives are the OoD tasks (lines and quadratic tasks) flagged as such; the false-positives are the in-distribution tasks (sines) flagged as OoD.

For the predictions, we plot the average and 95\% CI of the MSE on 1,000 tasks (100 queried inputs each).
In the \trainingrandom{} case, we also compute the average and 95\% CI on 5 different random projections trainings.

\section{Training and test details for the multi cluster case}
\label{app:train-details-multi}

\subsection{Training details of \approach{} (multi-cluster case)}
The model is a neural network with $2$ hidden layers, $40$ neurons each, with a ReLU non-linearity.
In our multi-cluster experiment, $\alpha=2$: the clusters consist of the sine cluster and the linear cluster (see Appendix~\ref{app:problem-details}).

For all the trainings, the training is performed with $\ntasksperepoch =24$ tasks per epoch (with an infinite number of available sine tasks and linear tasks during training \textit{ie} $\ntraintasks=+\infty$), and at each epoch the context inputs are randomly drawn from $[-5, 5]$ (which means that $\ntraindatapoints=+\infty$).
In accordance with our equal probability assumption from Section Methods, at each epoch $\ntasksperepoch/2=12$ tasks come from the sine cluster and $\ntasksperepoch/2=12$ tasks come from the linear cluster.

For all trainings, the number of context inputs seen during training is $\batchsize=10$.

For all trainings, we train \approach{} algorithm on $60,000$ epochs: because we deal with the \trainingfim{} case, we allow half of the epochs ($30,000$) to the training \textit{before} finding the intermediate $\paramlin$ and $\{\priormean_j\}_{j=1}^{j=\alpha}$ (see Algorithm~\ref{alg:meta_training_learnt_cov}), and the other half \textit{after}.

In the \trainingfim{} case, a finite number of tasks is needed to compute the FIM: thus we build an artificial finite dataset of $\ntraintasks=100$ and $\ntraindatapoints=\pdim$ (arbitrary, but chosen so that $\jac(\paramlin, \xallcontextinput) \jac(\paramlin, \xallcontextinput)^\top$ gets a chance to be full-rank \textit{ie} contain as much information as possible), that is used only for that computation.

For all trainings, the subspace size is $\sdim=10$.

For all trainings, the meta-optimizer is Adam with an initial learning rate of $0.001$.
The noise $\stdnoise=0.05$, equal to the noise added to the context data.
We compute the NLL using \citet{rasmussen}'s implementation.

When training with \trainingmixt, we make sure to initialize $\scaling_1$ and $\scaling_2$ randomly with $\normal(\zeromean,~0.5 \mI)$, so that the meta-learning can effectively differentiate the two clusters.
Also, in the \trainingmixt{} case, the mean $\priormean$ is unique \textit{before} computing the FIM.
Once we have found the projection directions, we initialize $(\priormean_1, \priormean_2)$ with the intermediate $\priormean$, thus yielding two Gaussians.

\subsection{Training details of MAML (multi-cluster case)}
We reimplement and train a MAML baseline to compare our results with that of MAML.
A large part of these hyperparameters is directly inspired from \citet{maml}'s work.

The model is a neural network with $2$ hidden layers, $40$ neurons each, with a ReLU non-linearity.
In our multi-cluster experiment, $\alpha=2$: the clusters consist of the sine cluster and the linear cluster (see Appendix~\ref{app:problem-details}).

The training is performed with $\ntasksperepoch=24$ tasks per epoch (with an infinite number of available sine and linear tasks during training \textit{ie} $\ntraintasks=+\infty$) and at each epoch the context and query inputs are randomly drawn from $[-5, 5]$.
In accordance with our equal probability assumption from Section Methods, at each epoch $\ntasksperepoch/2=12$ tasks come from the sine cluster and $\ntasksperepoch/2=12$ tasks come from the linear cluster.

For all trainings, the number of context datapoints is $\batchsize=10$, and the number of query datapoints is $L=10$.
We meta-train for $70,000$ epochs.

For all trainings, the meta-optimizer is Adam with an initial learning rate of $0.001$.
The inner learning-rate is kept constant, at $0.001$.
The number of inner updates is $5$ during training, and $10$ at test time.

\subsection{Training details of MMAML (multi-cluster case)}
We train a MMAML by using \citet{mmaml}'s code, using the best setting mentioned in their paper (FiLM).
We adapt their code to train it on our sine and linear clusters: we add an offset to their sine cluster (via their parameter \texttt{bias}), change the phase from $\sin(x - \varphi)$ to $\sin(x + \varphi)$ and specify via the arguments the slope range ($[-1, 1]$) and the y-intercept range ($[0, 0]$, because it does not vary in our case) of the line tasks.
We also change the number of context inputs that we set to $\batchsize=10$, to remain coherent with the rest of the trainings.
The rest of the hyperparameters are kept identical to the command specified in the repository of MMAML.
Finally, at test time, we make the query ground-truths noiseless, to remain coherent with the rest of the test conditions.

\subsection{Test details (multi-cluster case)}

For the OoD detection evaluation, we plot the AUC as a function of the number of context inputs $\batchsize$.
The AUC is computed using the NLL of the context inputs with respect to $\distparam$ (our uncertainty metric).
The true-positives are the OoD tasks (quadratic tasks) flagged as such; the false-positives are the in-distribution tasks (lines and sines) flagged as OoD.

For the predictions, we plot the average and 95\% CI of the MSE on 1,000 tasks (100 queried inputs each): half of them are sines and half of them are lines.

\section{Training and test details for the vision problem}
\label{app:train-details-vision}

The model is a deep neural network, identical to the one used by \citet{what-matters} except for the last layer: instead of doing a one-dimensional regression (where the output stands for an angle prediction), we perform a two-dimensional regression (where the output stands for a cosine and sine prediction).
This choice is motivated by the fact that the Gaussian noise assumed in Section Background cannot capture the complexity of an angle error (e.g., predicting $361^{\circ}$ should yield a low error when compared to the ground-truth angle $0^{\circ}$), but better renders the MSE that can be applied on cosine and sine.

\subsection{Training details of \approach{} (vision case)}
For all the trainings, there are $\ntasksperepoch=10$ tasks per epoch and $\batchsize=15$ context inputs per task.
When dealing with the \trainingfim{} case, we allow half of the epochs ($5,000$) to the training \textit{before} finding the intermediate $\paramlin$ and $\priormean$ (see Algorithm~\ref{alg:meta_training_learnt_cov}), and the other half after.

In the \trainingfim{} and \trainingrandom{} cases, the subspace size is $\sdim=100$.

For all the trainings, the meta-optimizer is Adam with an initial learning rate of $0.001$.
The noise is $\stdnoise=0.01$.
We compute the NLL using \citet{rasmussen}'s implementation.

\subsection{Training details of MAML (vision case)}
We train a MAML baseline to compare our results with that of MAML.
A large part of these hyperparameters is directly inspired from \citet{what-matters}.

The training is performed with $\ntasksperepoch=10$ tasks per epoch.
The number of context datapoints is $K=15$, and the number of query datapoints is $L=10$.
We meta-train for $50,000$ epochs.

For all trainings, the meta-optimizer is Adam with an initial learning rate of $0.0005$.
The inner learning-rate is kept constant, at $0.002$.
The number of inner updates is $5$ during training, and $20$ at test time.

\subsection{Test details (vision problem)}
At test time, we wrap our model with the $\arctan$ function to convert the predictions to angles.
Then, we compare the angle predictions with the ground-truth angles.
To do so, we use the following error from \cite{what-matters} to compare two angles:
\begin{displaymath}
    \mathcal{E}(\beta, \beta^*) = \min \{ \mathcal{E}_{\beta^+, \beta^*}, \mathcal{E}_{\beta, \beta^*}, \mathcal{E}_{\beta^-, \beta^*} \}
\end{displaymath}
where $\mathcal{E}_{\beta^\pm, \beta^*} = \vert y \pm 360 - y^* \vert$ and $\mathcal{E}_{\beta, \beta^*} = \vert y - y^* \vert$.

\begin{figure}
     \centering
     \begin{subfigure}[t]{0.45\textwidth}
         \centering
         \includegraphics[width=\textwidth]{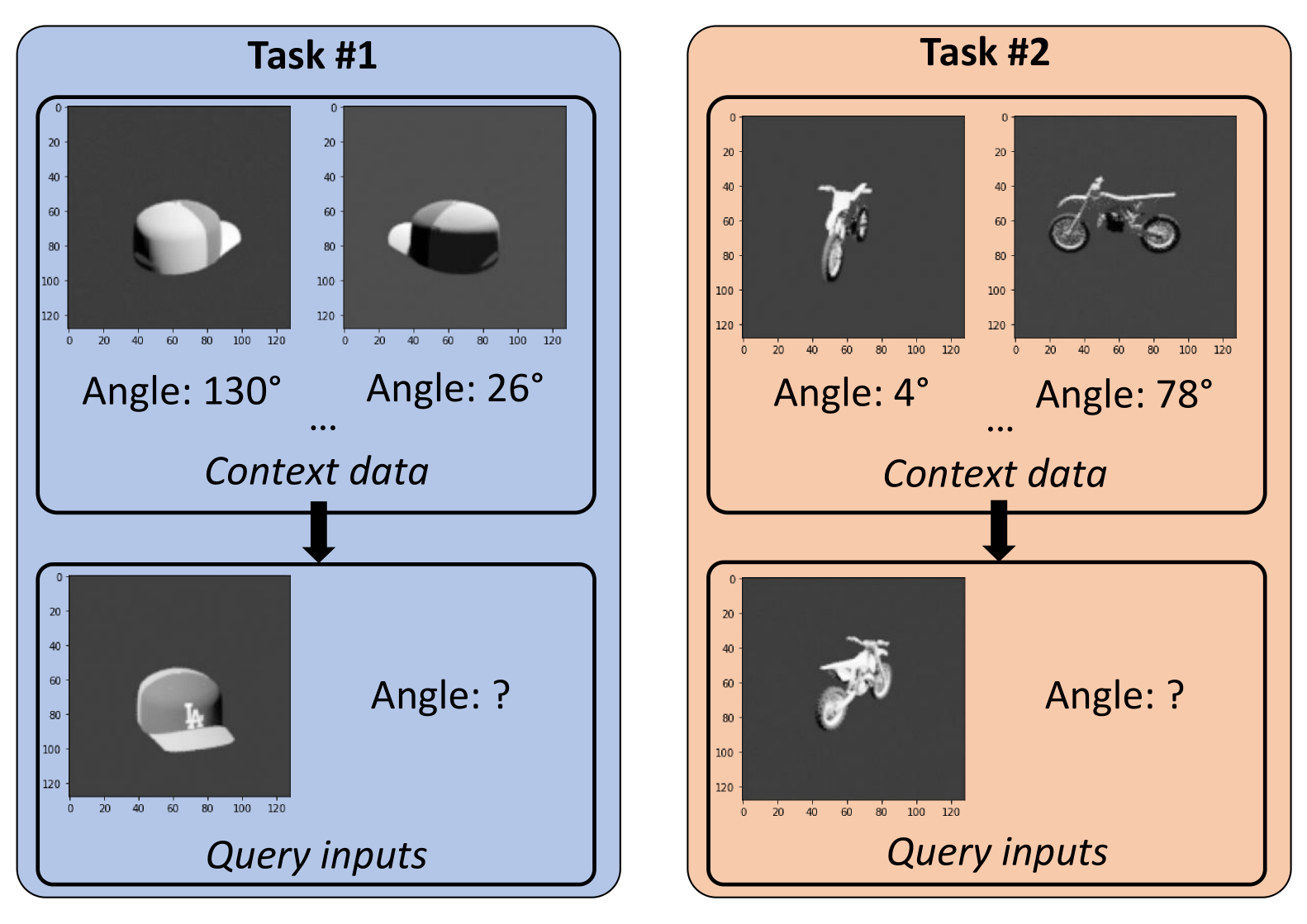}
         \caption{Vision Task}
     \end{subfigure}
     \begin{subfigure}[t]{0.45\textwidth}
         \centering
         \includegraphics[width=\textwidth]{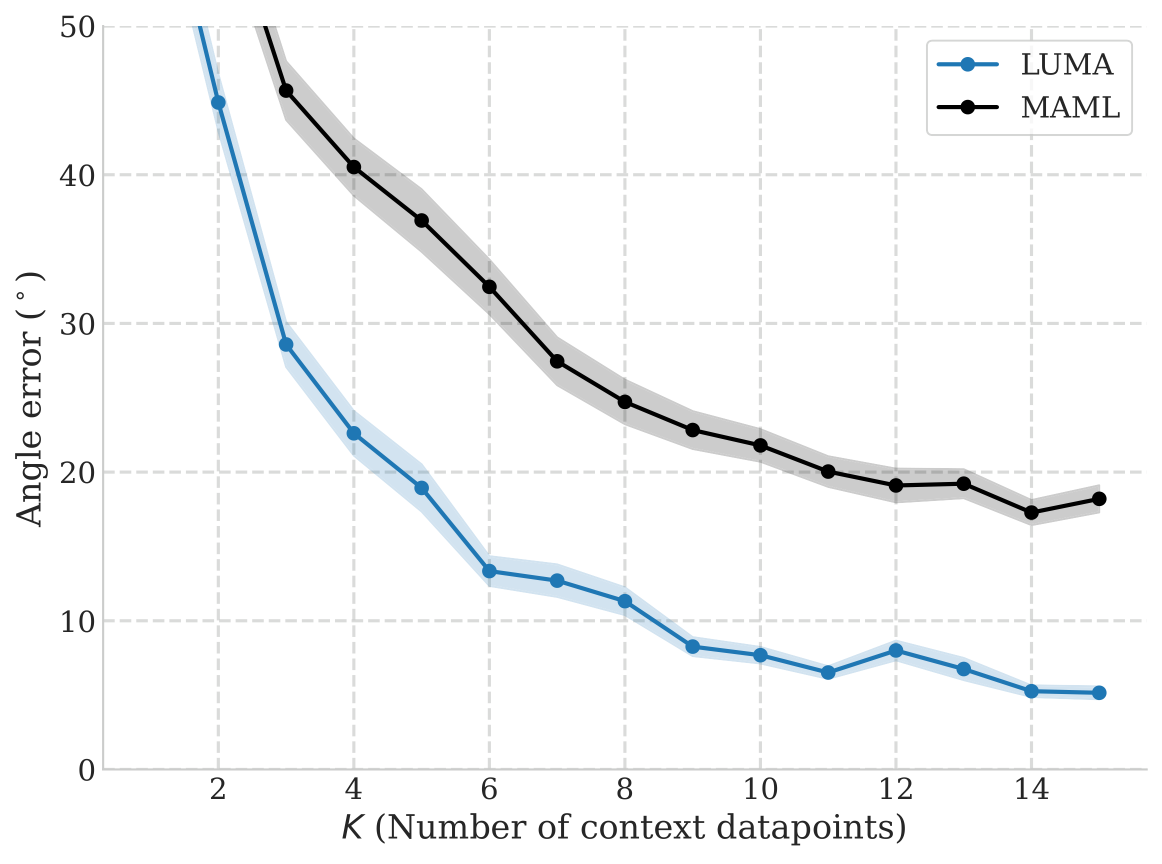}
         \caption{Angle Error}
     \end{subfigure}
    \caption{(a) Graphical depiction of the vision tasks. (b) Angle error with varying $\batchsize$.
    \approach{} provides more accurate predictions than the baseline.
       }
    \label{fig:vision-performance}
\end{figure}

When plotting the performance (Figure~\ref{fig:vision-performance}), we plot the average and 95\% CI on 100 tasks (15 queried inputs each). For \trainingrandom, the average and 95\% CI are computed on 5 different random projection trainings.

\begin{figure}[t]
     \centering
     \begin{subfigure}[t]{0.45\textwidth}
         \centering
         \includegraphics[width=\textwidth]{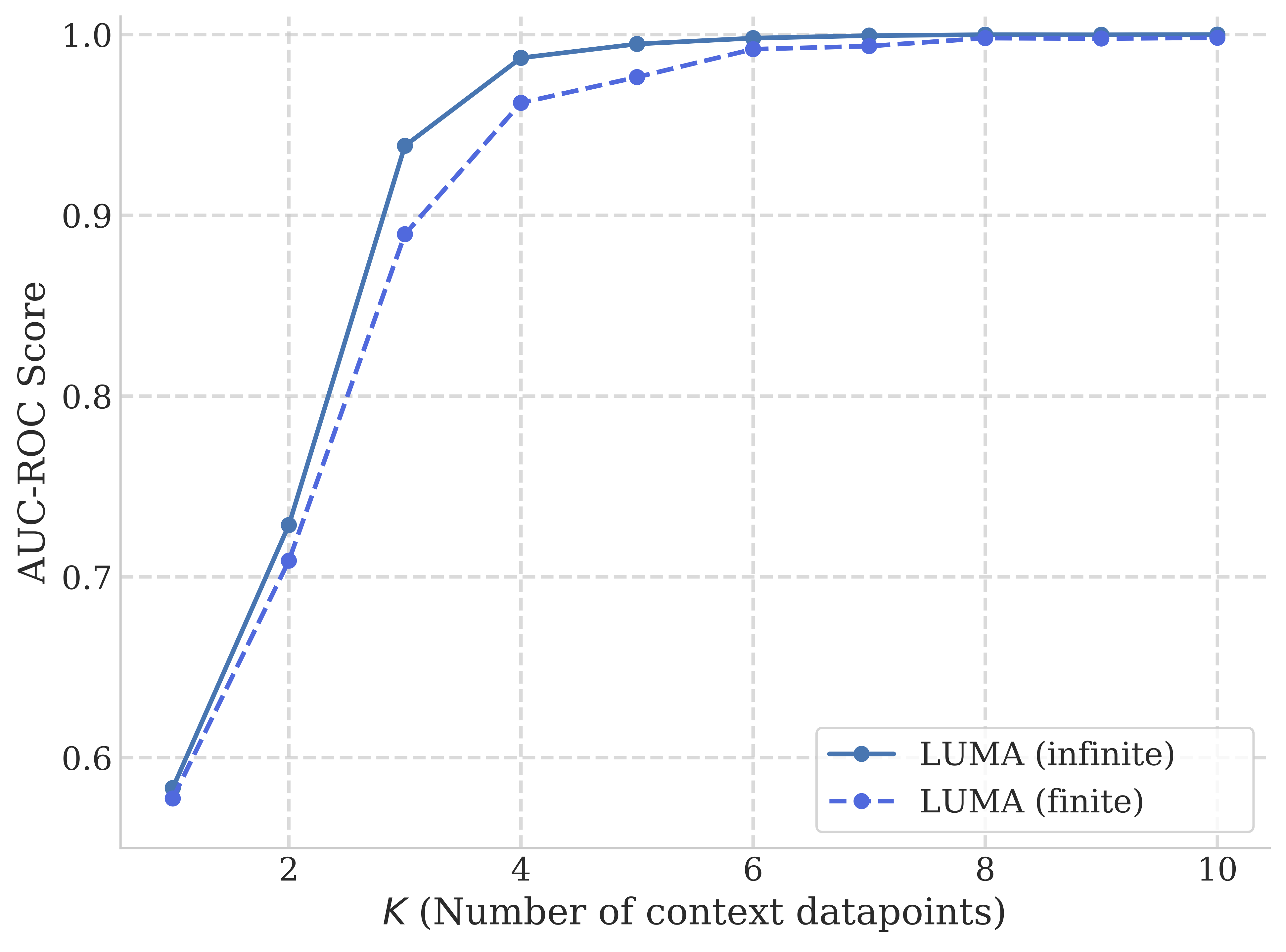}
         \caption{Unimodal}
         \label{fig:single-auc}
     \end{subfigure}
     \begin{subfigure}[t]{0.45\textwidth}
         \centering
         \includegraphics[width=\textwidth]{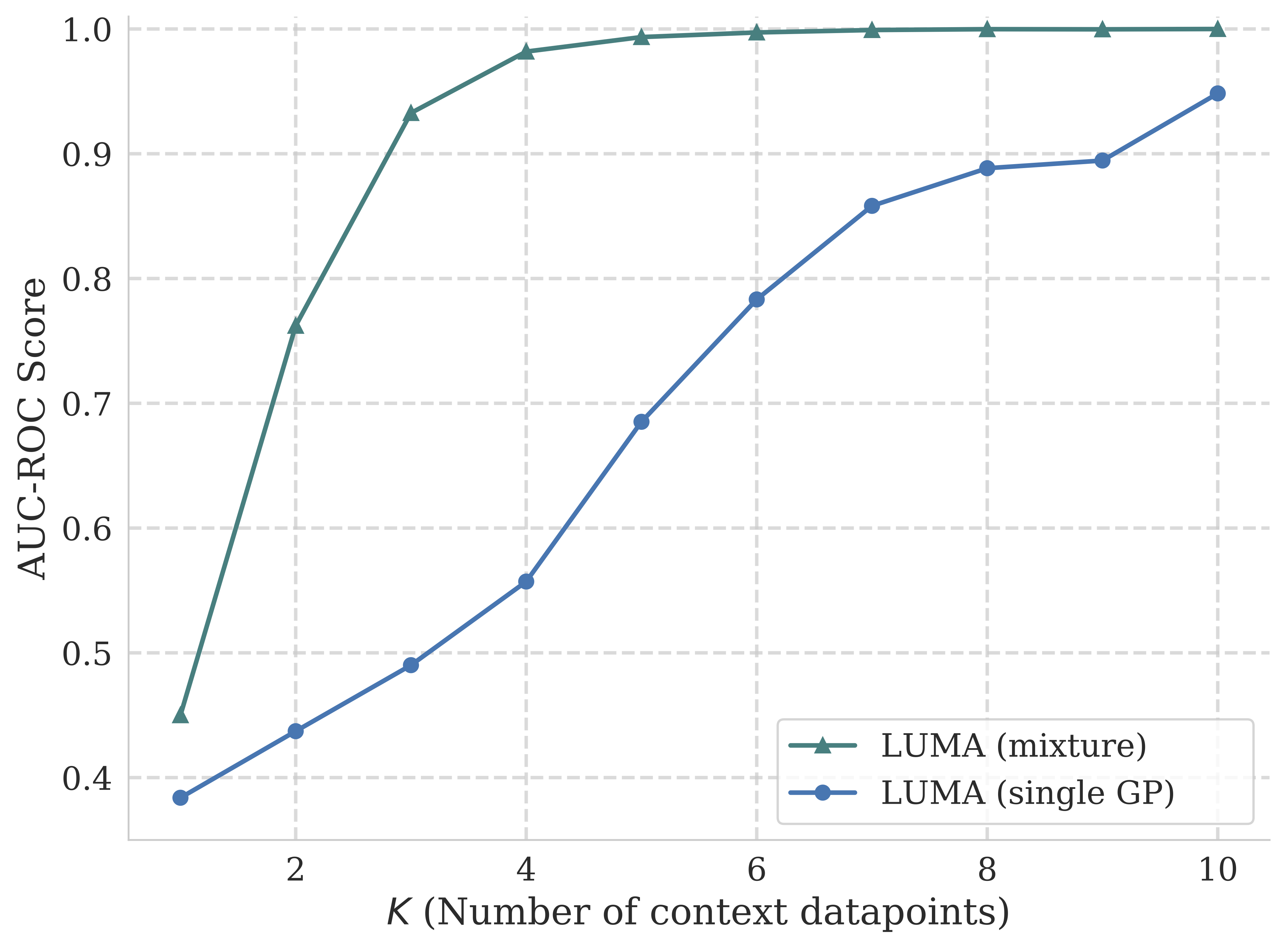}
         \caption{Multimodal}
         \label{fig:multi-auc}
     \end{subfigure}
    \caption{OoD detection performance:
    AUC-ROC score is evaluated with varying $\batchsize$.
    \approach{} achieves high accuracy even with a limited number of context datapoints, in both unimodal and multimodal settings. In the multimodal setting, the mixture model further improves performance, as expected.}
    \label{fig:ood-performance}
\end{figure}

\end{document}

%% file: math_commands.tex
\usepackage{amsmath,amsfonts,bm}

\def\eqref#1{equation~\ref{#1}}

\def\1{\bm{1}}

\def\vmu{{\bm{\mu}}}
\def\vtheta{{\bm{\theta}}}

\def\vs{{\bm{s}}}

\def\vx{{\bm{x}}}
\def\vy{{\bm{y}}}
\def\vz{{\bm{z}}}

\def\mF{{\bm{F}}}

\def\mI{{\bm{I}}}
\def\mJ{{\bm{J}}}

\def\mQ{{\bm{Q}}}

\def\mS{{\bm{S}}}

\def\mW{{\bm{W}}}
\def\mX{{\bm{X}}}
\def\mY{{\bm{Y}}}

\def\mSigma{{\bm{\Sigma}}}

\DeclareMathAlphabet{\mathsfit}{\encodingdefault}{\sfdefault}{m}{sl}
\SetMathAlphabet{\mathsfit}{bold}{\encodingdefault}{\sfdefault}{bx}{n}

\newcommand{\R}{\mathbb{R}}

\newcommand{\KL}{D_{\mathrm{KL}}}

%% file: preamble.tex
\usepackage{xcolor}
\newcommand{\revise}[1]{#1}

\newcommand{\modelsingleinput}{f}
\newcommand{\model}{g}

\newcommand{\pdim}{P}
\newcommand{\param}{\vtheta}
\newcommand{\paramlin}{\param_0}
\newcommand{\paramcorr}{\hat{\param}}
\newcommand{\parampert}{\Tilde{\param}}

\newcommand{\xdim}{{D_x}}
\newcommand{\xinput}{\mX}
\newcommand{\xsingleinput}{\vx}
\newcommand{\xcontextinput}{\xinput^i}
\newcommand{\xallcontextinput}{\widetilde{\xinput}^i}
\newcommand{\xqueryinput}{\xinput^i_*}

\newcommand{\ydim}{{D_y}}
\newcommand{\youtput}{\mY}
\newcommand{\yvectorizedoutput}{\vy}
\newcommand{\ysingleoutput}{\vy}
\newcommand{\ycontextoutput}{\youtput^i}
\newcommand{\yallcontextoutput}{\widetilde{\youtput}^i}
\newcommand{\yqueryoutput}{\youtput^i_*}

\newcommand{\batchsize}{K}
\newcommand{\ntraindatapoints}{M}

\newcommand{\sdim}{r}
\newcommand{\proj}{\mQ}
\newcommand{\Scaling}{\mS}
\newcommand{\scaling}{\vs}

\newcommand{\featuremap}{\phi}
\newcommand{\kernel}{k}

\newcommand{\noisecov}{\mSigma_\noise}
\newcommand{\stdnoise}{\sigma_\noise}
\newcommand{\noise}{\varepsilon}

\newcommand{\normal}{\mathcal{N}}
\newcommand{\zeromean}{\bm{0}}
\newcommand{\priormean}{\vmu}
\newcommand{\priorcov}{\mSigma}
\newcommand{\postpredmean}{\vmu_{\youtput \mid \xcontextinput}}
\newcommand{\postpredcov}{\mSigma_{\youtput \mid \xcontextinput}}
\newcommand{\meanpriorpred}{\vmu_{\text{prior}}}
\newcommand{\meanpostpred}{\vmu_{\text{post}}}
\newcommand{\covpriorpred}{\text{cov}_{\text{prior}}}
\newcommand{\covpostpred}{\text{cov}_{\text{post}}}

\newcommand{\task}{\mathcal{T}^i}

\newcommand{\dist}{p(f)}
\newcommand{\distparam}{\Tilde{p}_\xi(f)}

\newcommand{\gaussian}{\mathcal{G}}

\newcommand{\fim}{\mF}
\newcommand{\fimd}{\fim(\paramlin, \dataset)}
\newcommand{\dataset}{\mathcal{D}}
\newcommand{\ntraintasks}{N}
\newcommand{\ntasksperepoch}{n}

\newcommand{\diag}[1]{\text{diag} (#1)}

\newcommand{\sketch}{\mathcal{S}}
\newcommand{\sketchY}{\mY}
\newcommand{\sketchW}{\mW}
\newcommand{\sketchOm}{\bm{\Omega}}
\newcommand{\sketchPsi}{\bm{\Psi}}

\newcommand{\cluster}{\mathcal{C}}
\newcommand{\nclusters}{\alpha}

\newcommand{\jac}{\mJ}

\newcommand{\firstparam}{\bm{\lambda}}
\newcommand{\lastparam}{\bm{\rho}}
\newcommand{\lastdim}{{N_\psi}}

\newcommand{\approach}{\textsc{LUMA}}
\newcommand{\approachLong}{\textit{Low-rank Uncertainty-aware Meta-learning Algorithm}}

\newcommand{\trainingid}{\approach-\textsc{I}}
\newcommand{\trainingrandom}{\approach-\textsc{R}}
\newcommand{\trainingfim}{\approach-\textsc{F}}
\newcommand{\trainingmixt}{\textsc{Mixt}}

\DeclareMathOperator{\add}{sum}
\DeclareMathOperator{\vectorized}{vec}